\PassOptionsToPackage{table,dvipsnames}{xcolor}
\documentclass{article} 
\usepackage{iclr2026_conference,times}

\usepackage{amsmath,amsfonts,bm}

\def\eqref#1{equation~\ref{#1}}

\def\1{\bm{1}}

\DeclareMathAlphabet{\mathsfit}{\encodingdefault}{\sfdefault}{m}{sl}
\SetMathAlphabet{\mathsfit}{bold}{\encodingdefault}{\sfdefault}{bx}{n}

\usepackage{xurl}
\usepackage{hyperref}
\usepackage{xspace}
\usepackage{amsthm}
\usepackage{xcolor}
\usepackage{amssymb,mathrsfs,amsmath}
\usepackage{mathtools}

\usepackage{graphicx}
\usepackage{float}
\usepackage{placeins}
\usepackage{wrapfig}
\usepackage{subfigure}
\usepackage{caption}

\usepackage{booktabs}
\usepackage{multirow}
\usepackage{makecell}
\usepackage{tabularx}
\usepackage{colortbl}
\usepackage{arydshln}
\usepackage{threeparttable}

\usepackage[ruled,vlined,noend]{algorithm2e}
\usepackage[most]{tcolorbox}
\usepackage{enumitem}
\usepackage{bbding}
\usepackage{pifont}
\usepackage{bbm}
\usepackage{siunitx}

\newcommand{\llmname}[1]{\textsc{#1}}
\makeatletter\newcommand{\tightparagraph}{\@startsection{paragraph}{4}{\z@}{0.8ex plus 0.2ex minus 0.1ex}{-1em}{\normalsize\bf}}\makeatother
\makeatletter\newcommand{\baselineparagraph}{\@startsection{paragraph}{4}{\z@}{1.2ex plus 0.3ex minus 0.1ex}{-1em}{\normalsize\bf}}\makeatother

\makeatletter\newcommand{\methodparagraph}{\@startsection{paragraph}{4}{\z@}{1.8ex}{-1em}{\normalsize\bf}}\makeatother

\usepackage{listings}

\newcommand{\ourmethod}{{\textsc{LOPD}}\xspace}

\definecolor{bittersweet}{rgb}{1.0, 0.44, 0.37}
\definecolor{mygreen}{rgb}{0.29, 0.7, 0.48}
\usepackage{pifont}
\definecolor{demphcolor}{RGB}{144,144,144}

\definecolor{mygray}{gray}{0.4}
\definecolor{autopurple}{HTML}{7030A0}
\definecolor{dyna_yellow}{HTML}{BF9000}
\definecolor{adaptive_blue}{HTML}{0070C0}
\definecolor{darksalmon}{rgb}{0.91, 0.59, 0.48}
\definecolor{emerald}{rgb}{0.31, 0.78, 0.47}
\definecolor{green(pigment)}{rgb}{0.0, 0.65, 0.31}
\definecolor{amaranth}{rgb}{0.9, 0.17, 0.31}
\definecolor{iris}{rgb}{0.35, 0.31, 0.81}
\definecolor{uu}{rgb}{0.95, 0.51, 0.51}
\definecolor{spirodiscoball}{rgb}{0.06, 0.75, 0.99}
\usepackage{svg}

\definecolor{lopdteal}{HTML}{287866}
\newtcolorbox{insightbox}[1]{
  enhanced jigsaw,
  breakable,
  colback=lopdteal!4!white,
  colframe=lopdteal!82!black,
  boxrule=0.8pt,
  arc=4pt,
  left=7pt,
  right=7pt,
  top=6pt,
  bottom=4pt,
  before skip=5pt,
  after skip=5pt,
  fontupper=\small,
  title={#1},
  colbacktitle=lopdteal,
  coltitle=white,
  fonttitle=\scriptsize\bfseries,
  attach boxed title to top left={xshift=12pt,yshift=-2.6mm},
  boxed title style={
    colback=lopdteal,
    colframe=lopdteal,
    boxrule=0pt,
    arc=2pt,
    left=4pt,
    right=4pt,
    top=0.5pt,
    bottom=0.5pt
  }
}

\colorlet{bestgreen}{SeaGreen!50}
\colorlet{secondgreen}{SeaGreen!30}
\newcommand{\bestcell}[1]{\cellcolor{bestgreen}\textbf{#1}}
\newcommand{\secondcell}[1]{\cellcolor{secondgreen}#1}

\usepackage{fontawesome5}
\newcommand{\hficon}{\raisebox{-0.20\height}{\includegraphics[height=1.05em]{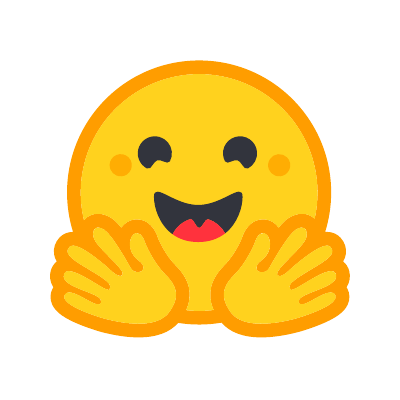}}}
\newcommand{\qwenicon}{\raisebox{-0.18\height}{\includegraphics[height=0.88em]{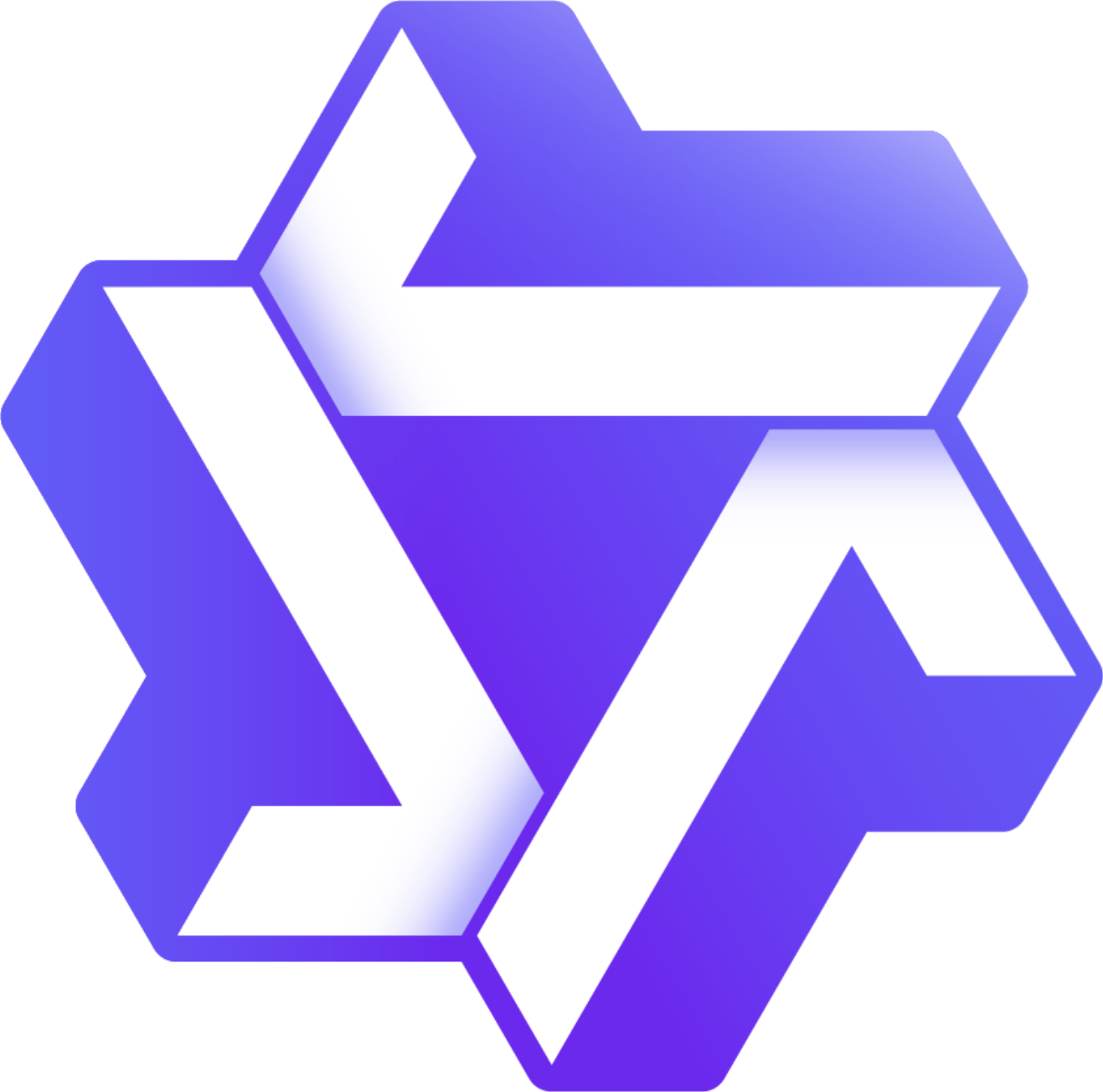}}}
\newcommand{\olmoicon}{\raisebox{-0.15\height}{\includegraphics[height=0.62em]{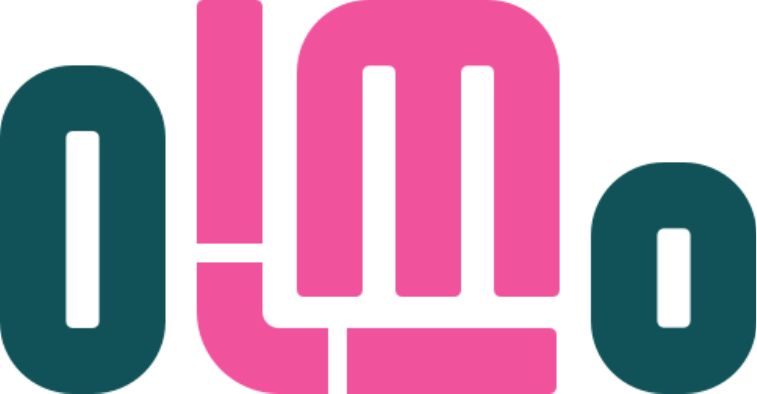}}}

\usepackage{cleveref}
\crefname{appendix}{Appendix}{Appendices}
\Crefname{appendix}{Appendix}{Appendices}
\SetKwInOut{Input}{Input}\SetKwInOut{Output}{Output}

\SetCommentSty{mycommfont}
\SetKwComment{Comment}{$\triangleright$ }{}

\hypersetup{
    colorlinks=true,
    linkcolor=RoyalBlue,
    citecolor=ForestGreen,
    filecolor=RoyalBlue,
    urlcolor=RoyalBlue,
    }

\title{Latent On-Policy Self-Distillation}

\author{Guibin Zhang$^{1,*}$, Jiayang Lyu$^{2,*}$, Ran Sun$^2$, Xinlei Yu$^{1}$, \\
\textbf{Haoyu Zhao}$^{1}$, \textbf{Qibing Ren}$^{3\dagger}$, \textbf{Shuicheng Yan}$^{1\dagger}$
\\[0.25em]
$^1$National University of Singapore\; $^2$Beijing University of Posts and Telecommunications \\
$^3$Shanghai Jiao Tong University $^\dagger$ Corresponding $^*$ Equal Contribution 
\\[0.05em]
\faGithub\ \textbf{GitHub:} \href{https://github.com/bingreeky/LOPD}{\texttt{github.com/bingreeky/LOPD}}\\[-0.02em]
\hficon\ \textbf{Hugging Face:}\quad
\qwenicon\,\href{https://huggingface.co/liunanfu1992/Qwen3-8B-LOPD}{\texttt{Qwen3-8B-LOPD}}\quad
\olmoicon\,\href{https://huggingface.co/liunanfu1992/OLMo-3-7B-Think-LOPD}{\texttt{Olmo3-7B-LOPD}}
}

\iclrfinalcopy 
\begin{document}

\maketitle
\lhead{Latent On-Policy Self-Distillation}

\begin{abstract}
Enabling agents to learn from experience and internalize it into their policy has become a central problem in self-evolving AI. On-policy self-distillation (OPSD) offers an effective pathway by using a privileged self-teacher to provide dense supervision on the student's own trajectories; however, existing methods still rely heavily on designer-specified privileged artifacts (\textit{e.g.}, answers, feedback, skills, or trajectories), limiting the end-to-end learnability and scalability required for continual self-improvement. In this work, we introduce \textbf{Latent On-Policy Self-Distillation} (\ourmethod), which, rather than proposing another hand-crafted OPSD variant with a newly prescribed form of privileged context, makes the teacher's privileged context itself learnable end-to-end from experience. Technically, \ourmethod retrieves relevant experiences and composes them into continuous latent tokens that condition a self-teacher, while the student generates trajectories from the task and interaction history and receives dense token-level supervision at every visited prefix. We further introduce a privileged-margin objective to stabilize and regulate the learning of latent context. Empirically, \ourmethod demonstrates \textbf{(I) strong performance}, outperforming RLVR and representative OPSD methods including OPSD, SDPO, and Skill-SD across both agentic tool use and code generation; and \textbf{(II) high learning efficiency}, surpassing GRPO and Skill-SD with less than $30\%$ of their rollout budget. Ablation studies further provide direct evidence that making privileged context learnable is necessary for realizing these gains. Together, these results position \ourmethod as a step toward a more scalable and self-directed paradigm for agent evolution.
\end{abstract}

\vspace{-0.4em}
\section{Introduction}

\emph{How should an agent learn from experience?}
A natural answer is to expose the model to prior successful trajectories, corrective feedback, or expert demonstrations, and then train it to internalize the behaviors they reveal.
This intuition has made on-policy distillation (OPD) a central paradigm for experience learning in large language models: the student first samples its own trajectory, and a teacher then provides dense token-level supervision on the states the student actually visits~\citep{song2026surveyonpolicydistillationlarge,li2026rethinkingonpolicydistillationlarge}.
Compared with off-policy imitation, OPD reduces the mismatch between training and inference; compared with reinforcement learning with verifiable rewards (RLVR), it converts sparse outcome feedback into fine-grained distributional signals over the student's own rollout~\citep{hbotter2026reinforcementlearningselfdistillation, yang2026selfdistilledrlvr}.
In this view, OPD is not merely a compression technique for a stronger model, but a mechanism for turning external experience into persistent policy improvement.

On-policy self-distillation (OPSD) has recently emerged as an especially appealing form of this paradigm because it removes the dependence on a separate, stronger teacher.
Instead, the teacher and the student are instantiated from the same model under different contexts: the student acts from the task and interaction history, while the teacher additionally receives privileged information such as verified reasoning traces, final answers, environment feedback, or successful prior rollouts~\citep{zhao2026selfdistilledreasoneronpolicyselfdistillation,hbotter2026reinforcementlearningselfdistillation,penaloza2026privilegedinformationdistillation}.
The privileged context makes the teacher distribution more informative than the student's, enabling dense feedback without querying an external model.
This design shifts the central question of OPSD from \emph{which teacher is strong enough?} to \emph{what privileged context should be given to the self-teacher?}
The answer matters: the context must expose useful experience, make token-level supervision sharper, and yet remain compatible with a student that will not see that context at inference time.

\begin{figure}
    \centering
    \includegraphics[width=1.0\linewidth]{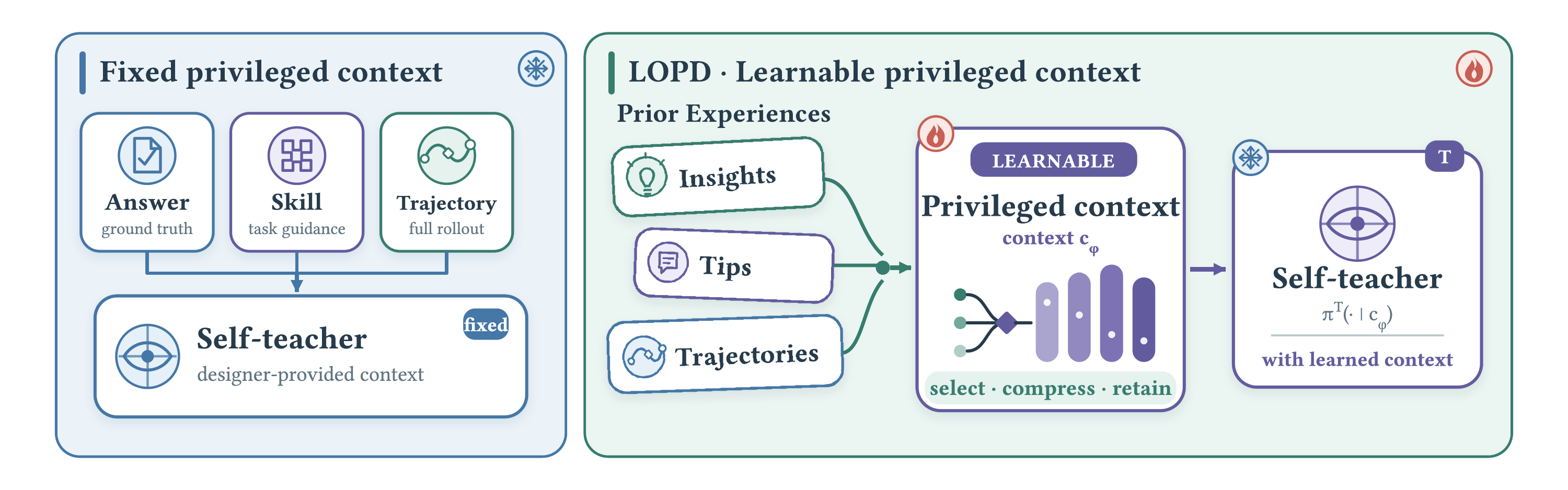}
    \vspace{-1.8em}
    \caption{
    \textbf{From fixed to learnable privileged context.}
    Existing OPSD methods condition the self-teacher on designer-specified answers, skills, or trajectories. \ourmethod instead learns latent context from prior experience and distills its dense supervision into the same on-policy student.
    Insights/tips illustrate compatible sources; \ourmethod instantiates the substrate using trajectories only.
    }
    \label{fig:privileged-context-paradigms}
    \vspace{-0.3em}
\end{figure}

Existing substrates for privileged experience remain imperfect in a more fundamental sense.
Although prior work instantiates privileged context in many forms---verified answers, reasoning traces, environment feedback, contrastive evidence, or action-only trajectories~\citep{zhao2026selfdistilledreasoneronpolicyselfdistillation,hbotter2026reinforcementlearningselfdistillation,yu2026dopd,penaloza2026privilegedinformationdistillation,yang2026selfdistilledrlvr}---these contexts are typically hand-designed or rule-extracted transformations of experience.
They decide \emph{a priori} what the teacher should see: an oracle answer, a textual reflection, a successful rollout, a retrieved document, or another discrete artifact.
Such substrates can be useful, but they also constrain self-distillation to the information format chosen by the designer, rather than allowing the teacher to learn which aspects of prior experience are actually useful for supervising the student's current trajectory.
This motivates a substrate-level question for OPSD:
\begin{insightbox}{Research Question}
\emph{Can privileged context itself be learned end-to-end from experience, so that the self-teacher, rather than a designer-specified rule, determines what experiential knowledge to retain and how to encode it for dense on-policy supervision?}
\end{insightbox}

To address this challenge, we propose \textbf{Latent On-Policy Self-Distillation} (\ourmethod), a framework that realizes privileged context as a \emph{learnable latent substrate}.
Its training pipeline otherwise follows standard OPSD: the student first rolls out trajectories from the task and interaction history, and the teacher re-evaluates every visited prefix to provide dense token-level distributions, against which the student is optimized through reverse-KL distillation.
The key difference is that the teacher is conditioned not on a pre-defined privileged artifact, but on learnable latent context instantiated by a composer that transforms retrieved experiences into compact continuous tokens.
Because this context is differentiable, the same training process that improves the student also teaches the composer which aspects of experience to retain and how to organize them into effective teacher supervision.
To ensure that this learning makes the teacher more informative rather than merely easier for the student to match, we introduce a privileged-margin constraint that requires the teacher to maintain a verifiable log-probability advantage over the student.
After training, only the student is retained: inference requires no experience database, retrieval module, composer, or latent context.
In this way, \ourmethod turns experience from a hand-crafted input artifact into an end-to-end learnable supervision substrate whose benefits are internalized by the student policy.

\begin{insightbox}{Positioning}
This work does not aim to introduce yet another elaborate OPSD variant by hand-designing a new form of privileged context. Its central claim is at the level of the \emph{experience substrate}: if self-evolving AI is to scale, neither the useful content of experience nor its representation should be fixed by designer heuristics. A scalable system should learn end-to-end what evidence to preserve and how to transform it into supervision. We study OPSD as a concrete mechanism for realizing this broader principle.
\end{insightbox}

Our contributions are summarized as follows:
\vspace{-0.5em}
\begin{itemize}[leftmargin=*, itemsep=0pt, parsep=0pt]
    \item \textbf{Context Re-formulation.}
    We reformulate privileged context from pre-defined, hand-designed textual artifacts into an end-to-end learnable latent substrate, allowing the teacher to automatically extract task-relevant supervision signals from prior experience.
    \item \textbf{Proposed Solution.}
    We develop \ourmethod, which composes retrieved experiences into continuous latent context for the self-teacher and jointly optimizes the privileged context and student through on-policy distillation. A privileged-margin constraint keeps the learned context informative and prevents the teacher from collapsing toward the student.
    \item \textbf{Experimental Validation.}   \ourmethod consistently surpasses RLVR and representative OPSD methods over three model backbones and seven benchmarks, and outperforms GRPO and Skill-SD with less than $30\%$ of their rollout budget. Further ablation studies directly establish the necessity of jointly learning the privileged context.

\end{itemize}

\vspace{-0.4em}
\section{Related Work}

\vspace{-0.6em}
\paragraph{On-Policy (Self-)Distillation.}
On-policy distillation trains a student on trajectories sampled from its own policy while querying a teacher for dense token-level supervision on the visited states~\citep{song2026surveyonpolicydistillationlarge,li2026rethinkingonpolicydistillationlarge}.
Recent work has applied this recipe broadly: reasoning and efficient post-training~\citep{2604.13010,2603.07079,2603.25562}, long-context modeling~\citep{2604.17535}, context and knowledge distillation~\citep{ye2026onpolicycontextdistillationlanguage,edgeopd2026internalizingprivilegedcontext}, GUI grounding~\citep{2605.00642}, and multimodal domains such as video grounding~\citep{2602.02994}, speech alignment~\citep{2603.24596}, and visual reasoning~\citep{2605.18740,2605.21924}.
Within this landscape, on-policy self-distillation (OPSD) removes the external teacher by instantiating teacher and student from the same model under different contexts: the student acts under the deployable input, while the teacher receives privileged context~\citep{zhao2026selfdistilledreasoneronpolicyselfdistillation,hbotter2026reinforcementlearningselfdistillation,penaloza2026privilegedinformationdistillation,yang2026selfdistilledrlvr,yu2026dopd}.
This line is best understood by the form of privileged context it supplies: verified answers and reasoning traces~\citep{zhao2026selfdistilledreasoneronpolicyselfdistillation,yang2026selfdistilledrlvr}, rich environment feedback or self-revision signals~\citep{hbotter2026reinforcementlearningselfdistillation,2604.12002,2605.12741}, contrastive or peer-rollout evidence~\citep{2605.12652,2606.11709}, action-only frontier-agent trajectories~\citep{penaloza2026privilegedinformationdistillation}, and retrieved or evidence-guided context~\citep{ye2026onpolicycontextdistillationlanguage,edgeopd2026internalizingprivilegedcontext}.
These designs demonstrate that privileged context is the key interface through which OPSD converts experience into supervision, but the context itself is usually pre-defined as a textual or discrete artifact; in contrast, \ourmethod makes the privileged context a learnable latent substrate jointly optimized with distillation.

\vspace{-0.6em}
\paragraph{Latent Computation.}
Latent computation uses continuous latent tokens/embeddings or hidden states as the carrier of LLM computation rather than natural language~\citep{yu2026latentspacefoundationevolution,zhu2025surveylatentreasoning}.
In \textit{reasoning}, latent tokens can expand the model's internal compute budget~\citep{deng2026llmlatentreasoningchain,hao2025traininglargelanguagemodels,amos2026latentreasoningsupervisedthinking}, letting the model perform deeper deliberation before producing an answer.
In \textit{memory}, latent states can serve as compact carriers of procedural~\citep{zhang2025memgenweavinggenerativelatent,yu2026vismemlatentvisionmemory}, factual~\citep{wang2024memoryllmselfupdatablelargelanguage,feng2026elasticmemlatentmemorylearnable}, and experiential~\citep{hou2026flashmemdistillingintrinsiclatent,wu2026tokmemonetokenproceduralmemory} information, preserving reusable experience without exposing long textual traces in the prompt.
In \textit{planning}, latent computation can represent intermediate plans, subgoals, or world-state summaries that guide downstream actions while remaining flexible and differentiable.
Our method is conceptually close to latent memory, but uses it in a different role: the latent state is not an inference-time augmentation, but a learnable privileged context through which an on-policy teacher supervises a student that does not observe this context.

\begin{figure}
    \centering
    \includegraphics[width=1\linewidth]{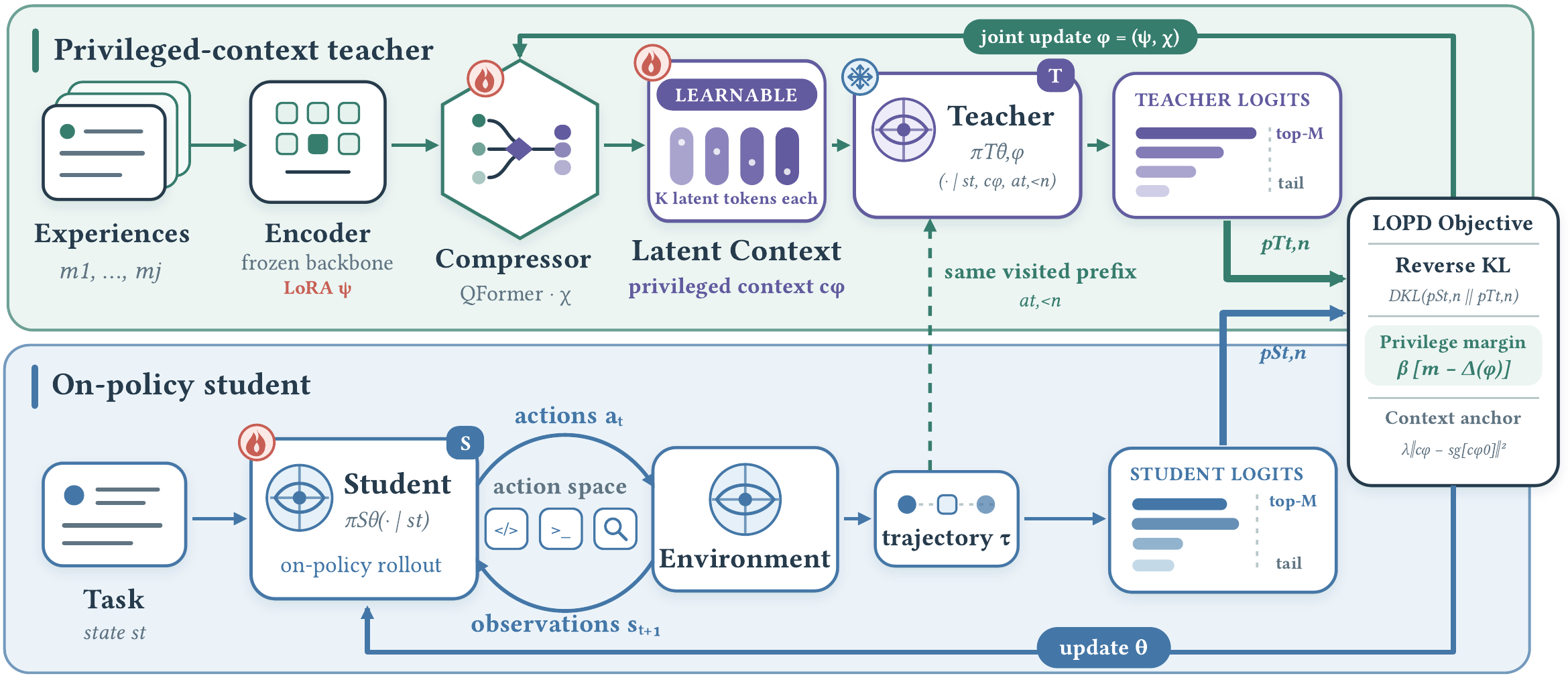}
    \vspace{-1.2em}
    \caption{
    \textbf{Overview of \ourmethod.}
    The student generates on-policy trajectories from the task and interaction history. A fixed-backbone teacher re-evaluates the same prefixes with latent context composed from retrieved experiences, and reverse-KL distillation matches their top-$M$-plus-tail distributions.
    }
    \label{fig:lopd-framework}
\end{figure}

\vspace{-0.4em}
\section{Method}
\vspace{-0.6em}
\subsection{Problem Setup: Privileged Context in OPSD}
\vspace{-0.6em}
We consider a multi-turn agent that receives a task $x$ and interacts with an environment through observations and actions. We call the acting policy the \emph{student}. At turn $t$, the student observes $\boldsymbol{s}_t=(x,\boldsymbol{o}_{\le t},\boldsymbol{a}_{<t})$, where $\boldsymbol{o}_{\le t}$ denotes observations so far and $\boldsymbol{a}_{<t}$ denotes previous actions, and samples actions to form a trajectory:
\begin{equation}
\boldsymbol{a}_t \sim \pi_\theta^{S}(\cdot\mid \boldsymbol{s}_t),\qquad
\boldsymbol{\tau}=(\boldsymbol{s}_1,\boldsymbol{a}_1,\ldots,\boldsymbol{s}_{|\boldsymbol{\tau}|},\boldsymbol{a}_{|\boldsymbol{\tau}|})\sim \pi_\theta^{S}(\cdot\mid x),
\label{eq:student-agent-rollout}
\end{equation}
where $\boldsymbol{a}_t=(a_{t,1},\ldots,a_{t,L_t})$ is the tokenized action at turn $t$, $L_t$ is its length, and $|\boldsymbol{\tau}|$ is the number of turns.
OPSD constructs a teacher by giving the same model additional privileged context. Abstractly, this context is obtained from some experience source $\mathcal{E}$ by a fixed transformation:
\begin{equation}
\boldsymbol{c}_{\mathrm{fix}}=\Phi_{\mathrm{fix}}(x,\mathcal{E}),\qquad
\pi_\theta^{T}(\cdot\mid \boldsymbol{s}_t,\boldsymbol{c}_{\mathrm{fix}})=\pi_\theta(\cdot\mid [\boldsymbol{s}_t;\boldsymbol{c}_{\mathrm{fix}}]),
\label{eq:fixed-privileged-context}
\end{equation}
where $\mathcal{E}$ may contain answers, traces, feedback, demonstrations, or retrieved trajectories, and $\Phi_{\mathrm{fix}}$ is a designer-specified rule that decides what artifact the teacher sees. Given a student trajectory, OPSD matches teacher and student distributions on the same visited prefixes:
\begin{equation}
\begin{aligned}
\mathcal{L}_{\mathrm{OPSD}}(\theta)
=
\mathbb{E}_{\boldsymbol{\tau}\sim\pi_\theta^{S}}
\left[
\sum_{t=1}^{|\boldsymbol{\tau}|}\sum_{n=1}^{L_t}
D\!\left(
\operatorname{sg}\!\left[\pi_\theta^{T}(\cdot\mid \boldsymbol{s}_t,\boldsymbol{c}_{\mathrm{fix}},a_{t,<n})\right]
\middle\|
\pi_\theta^{S}(\cdot\mid \boldsymbol{s}_t,a_{t,<n})
\right)
\right],
\end{aligned}
\label{eq:generic-opsd}
\end{equation}
where $a_{t,<n}$ is the action prefix before token $n$, $D$ is a token-level divergence, and $\operatorname{sg}[\cdot]$ denotes stop-gradient. This setup makes the limitation explicit: the supervision quality is bounded by a pre-defined context constructor. We instead parameterize the constructor:
\begin{equation}
\boldsymbol{c}_{\phi}=\Phi_\phi(x,\mathcal{E})=\langle e_1\rangle\oplus\langle e_2\rangle\oplus\cdots\oplus\langle e_K\rangle,
\label{eq:learnable-context}
\end{equation}
where $\Phi_\phi$ is a learnable composer, $\langle e_i\rangle$ is a continuous latent token, $K$ is the number of latent tokens per retrieved experience, and $\oplus$ denotes context concatenation. The goal of \ourmethod is to make privileged context itself learnable while preserving the defining asymmetry of OPSD: the teacher additionally observes $\boldsymbol{c}_\phi$, whereas the student continues to condition only on $\boldsymbol{s}_t$.

\vspace{-0.6em}
\subsection{Learnable Latent Privileged Context}\label{sec:learnablelatent}
\vspace{-0.6em}
\ourmethod instantiates $\Phi_\phi$ as a latent-context composer. We deliberately begin with a minimal experience pipeline. Offline, we retain successful rollouts in an experience bank, with each entry storing its task description and a compact action--result trace that omits verbose observations. For a current task $x$, a dense retriever embeds the query and each stored task--trajectory pair, then returns the top-$J$ entries under cosine similarity; these entries form $\mathcal{E}=\{m_j\}_{j=1}^{J}$. Full construction and retrieval details are provided in \Cref{app:retrieval}.

\begin{insightbox}{Raw Experience as a Minimal Substrate}
Our experience bank and similarity retriever are intentionally simple. Richer sources such as off-the-shelf agent skills, reusable codebooks, or learned retrievers can be incorporated without changing the framework. The point is not to prescribe another experience format, but to show that useful privileged context can be learned directly from minimally processed trajectories rather than hand-crafted rules.
\end{insightbox}

Given $x$ and the retrieved set $\mathcal{E}$, the composer first encodes each item into hidden states:
\begin{equation}
\mathbf{H}_j=\operatorname{Enc}_{\psi}(x,m_j)\in \mathbb{R}^{T_j\times d},
\label{eq:experience-encoder}
\end{equation}
where $m_j$ is the $j$-th retrieved trajectory, $\operatorname{Enc}_{\psi}$ is the encoder that maps the task--experience pair into hidden states, $T_j$ is the encoded sequence length, and $d$ is the hidden dimension.
The variable-length states are compressed into a fixed number of latent tokens by a lightweight latent compressor. In our implementation, this compressor is QFormer-style cross-attention with learned queries, although other latent-token generators can be used under the same abstraction:
\begin{equation}
\mathbf{E}_j=\operatorname{Comp}_{\chi}(\mathbf{H}_j;\mathbf{Q}_{\chi})=[\langle e_{j,1}\rangle,\ldots,\langle e_{j,K}\rangle]\in \mathbb{R}^{K\times d},
\label{eq:latent-composer}
\end{equation}
where $\operatorname{Comp}_{\chi}$ is the compressor, $\mathbf{Q}_{\chi}\in\mathbb{R}^{K\times d}$ is a learned query bank, and $\mathbf{E}_j$ is the latent representation of experience item $m_j$. We use $\phi\coloneqq(\psi,\chi)$ to denote all trainable composer parameters, comprising the encoder LoRA parameters $\psi$ and compressor parameters $\chi$; the encoder backbone itself remains frozen. Note that we do not treat the particular attention block as a contribution; its role is simply to produce a compact, differentiable context (details in \Cref{app:composer-arch}).
The teacher receives the latent tokens as ordinary context positions:
\begin{equation}
\boldsymbol{c}_{\phi}(x,\mathcal{E})=\bigoplus_{j=1}^{J}(\langle e_{j,1}\rangle\oplus\cdots\oplus\langle e_{j,K}\rangle),\qquad
\pi_{\bar{\theta},\phi}^{T}(\cdot\mid \boldsymbol{s},\boldsymbol{c}_{\phi})=\pi_{\bar{\theta}}(\cdot\mid [\boldsymbol{s};\boldsymbol{c}_{\phi}]),
\label{eq:latent-context-concat}
\end{equation}
where $\boldsymbol{s}$ is an interaction state defined above and $\boldsymbol{c}_{\phi}$ is the learned privileged context. Each $\langle e_{j,k}\rangle$ is implemented as a continuous embedding and behaves like a special token in the teacher's context window.
The composer is cold-started on successful trajectories with the backbone frozen, yielding $\phi_0$. During subsequent joint optimization, both the encoder LoRA parameters $\psi$ and compressor parameters $\chi$ remain trainable as part of $\phi$, while the teacher backbone $\bar{\theta}$ stays fixed (details in \Cref{app:cold-start}).

\vspace{-0.4em}
\subsection{Latent On-Policy Self-Distillation}
\vspace{-0.4em}

Having specified how the teacher is constructed, we now describe the self-distillation process. The student first produces a multi-turn trajectory conditioned only on $\boldsymbol{s}_t$. The composer then constructs $\boldsymbol{c}_{\phi}$ from retrieved experience, and the teacher re-evaluates the same student prefixes with this additional context. We instantiate the teacher as a frozen reference copy $\bar{\theta}$ initialized from the same backbone as the student; the two share architecture and initialization, but not weights after the student begins updating. For every turn $t$ and token position $n$, we define:
\begin{equation}
\begin{aligned}
\boldsymbol{p}^{S}_{t,n}=\pi_{\theta}^{S}\big(\cdot\mid \boldsymbol{s}_t,a_{t,<n}\big),\;
\boldsymbol{p}^{T}_{t,n}=\pi_{\bar{\theta},\phi}^{T}\big(\cdot\mid \boldsymbol{s}_t,\boldsymbol{c}_{\phi}(x,\mathcal{E}),a_{t,<n}\big),
\end{aligned}
\label{eq:multi-turn-teacher-student}
\end{equation}
where $\boldsymbol{p}^{S}_{t,n}$ and $\boldsymbol{p}^{T}_{t,n}$ are next-token distributions over the vocabulary $\mathcal{V}$, and $\bar{\theta}$ denotes a fixed base-model checkpoint whose parameters receive no gradient updates; however, the teacher's forward activations remain in the computation graph, so $\boldsymbol{p}^{T}_{t,n}$ is differentiable with respect to $\boldsymbol{c}_{\phi}$ and hence to $\phi$.
To keep logit-level distillation efficient, we follow~\citep{zhao2026selfdistilledreasoneronpolicyselfdistillation} and use the teacher top-$M$ vocabulary entries plus a tail bucket. This preserves the teacher's high-probability modes while accounting for the remaining probability mass:
\begin{equation}
\tilde{\boldsymbol{p}}^{T}_{t,n}(v)=
\begin{cases}
\boldsymbol{p}^{T}_{t,n}(v), & v\in \mathcal{V}_{t,n}^{M},\\
1-\sum_{u\in\mathcal{V}_{t,n}^{M}}\boldsymbol{p}^{T}_{t,n}(u), & v=\bot,
\end{cases}
\;
\tilde{\boldsymbol{p}}^{S}_{t,n}(v)=
\begin{cases}
\boldsymbol{p}^{S}_{t,n}(v), & v\in \mathcal{V}_{t,n}^{M},\\
1-\sum_{u\in\mathcal{V}_{t,n}^{M}}\boldsymbol{p}^{S}_{t,n}(u), & v=\bot,
\end{cases},
\label{eq:topk-tail}
\end{equation}
where $\mathcal{V}_{t,n}^{M}$ is the teacher top-$M$ support at prefix $(t,n)$ and $\bot$ denotes the aggregated tail event.
The distillation objective matches teacher and student on the student's own multi-turn trajectory:
\begin{equation}
\begin{aligned}
\mathcal{L}_{\mathrm{distill}}(\theta,\phi)
=
\mathbb{E}_{x\sim\mathcal{D}}\mathbb{E}_{\boldsymbol{\tau}\sim\pi_\theta^{S}(\cdot\mid x)}
\left[
\frac{
\sum_{t=1}^{|\boldsymbol{\tau}|}\sum_{n=1}^{L_t}
\omega_{t,n}\,
D_{\mathrm{KL}}\!\left(\tilde{\boldsymbol{p}}^{S}_{t,n}\middle\|\tilde{\boldsymbol{p}}^{T}_{t,n}\right)}
{\sum_{t=1}^{|\boldsymbol{\tau}|}\sum_{n=1}^{L_t}\omega_{t,n}}
\right],
\end{aligned}
\label{eq:lopd-distill}
\end{equation}
where $\mathcal{D}$ is the task distribution and $\omega_{t,n}\in\{0,1\}$ masks supervised action tokens. We use reverse KL so that the student concentrates on teacher-supported behavior. Because $\bar{\theta}$ denotes a fixed checkpoint while $\phi$ remains trainable, the distillation gradient flows through the frozen network back to the latent input (injection mechanism in \Cref{app:latent-injection}). This allows the composer to learn which experience features produce effective teacher supervision for the current student trajectory.

\vspace{-0.4em}
\paragraph{Privileged-Margin Constraint.} The distillation loss alone does not ensure that the teacher provides more effective supervision: an unconstrained composer can instead minimize \Cref{eq:lopd-distill} by moving $\boldsymbol{p}^{T}$ toward $\boldsymbol{p}^{S}$, yielding uninformative latent context without improving the student. We therefore prevent this collapse in two ways. \ding{182} First, the cold start in \Cref{sec:learnablelatent} initializes the composer to transform experience into informative latent context. \ding{183 }Second, we introduce outcome reward into the supervision signal, and more concretely, impose a privileged-margin constraint that ties the teacher's token-level advantage to the verified outcome of the complete trajectory. For each supervised token, we define the per-token privilege:
\begin{equation}
\delta_{t,n}(\phi)=\log\pi^{T}_{\bar{\theta},\phi}\!\left(a_{t,n}\mid \boldsymbol{s}_t,\boldsymbol{c}_{\phi},a_{t,<n}\right)-\operatorname{sg}\!\left[\log\pi^{S}_{\theta}\!\left(a_{t,n}\mid \boldsymbol{s}_t,a_{t,<n}\right)\right],
\label{eq:per-token-privilege}
\end{equation}
where $a_{t,n}$ is the student's sampled token and $\operatorname{sg}$ blocks gradients to $\theta$ through this path. Both log-probabilities are already computed during the teacher and student forward passes; $\delta_{t,n}$ requires only a gather, not an additional forward. We then use the trajectory-level verification signal $A(\boldsymbol{\tau})=2r(\boldsymbol{\tau})-1\in[-1,1]$, where $r(\boldsymbol{\tau})$ is the outcome reward, to favor teacher advantages on successful trajectories and suppress them on unsuccessful ones:
\begin{equation}
\Delta(\phi)=\mathbb{E}_{\boldsymbol{\tau}}\!\left[\frac{\sum_{t,n}\omega_{t,n}\,A(\boldsymbol{\tau})\,\delta_{t,n}(\phi)}{\sum_{t,n}\omega_{t,n}}\right].
\label{eq:privileged-margin}
\end{equation}
The full \ourmethod objective constrains $\phi$ to maintain a minimum privilege level $m>0$:
\begin{equation}
\min_{\theta,\phi}\;\max_{\beta\ge 0}\;\;\mathcal{L}_{\mathrm{distill}}(\theta,\phi)\;+\;\beta\!\left(m-\Delta(\phi)\right)+\;\lambda\left\|\boldsymbol{c}_{\phi}-\operatorname{sg}\!\left[\boldsymbol{c}_{\phi_0}\right]\right\|_2^2,
\label{eq:lopd-objective}
\end{equation}
where $\beta$ is a dual variable updated by $\beta\leftarrow[\beta+\eta_\beta(m-\Delta(\phi))]_+$, and the anchor term penalizes drift from the initialization $\phi_0$ in latent space, requiring no additional forward pass. If the composer degenerates to uninformative context, $\pi^{T}\to\pi^{S}$, $\delta_{t,n}\to 0$, $\Delta\to 0<m$, and the dual penalty activates---structurally excluding the trivial solution. Within the feasible region, outcome-weighted privilege steers the composer toward evidence that supports successful behavior, while the distillation gradient determines how that evidence is selected and encoded.

\begin{algorithm}[!t]
\DontPrintSemicolon
\SetAlgoLined
\SetAlgoNoEnd
\SetAlgoNlRelativeSize{-1}
\SetAlgoRefRelativeSize{-1}
\LinesNumbered
\Input{Task distribution $\mathcal{D}$; retriever $\operatorname{Ret}$; student $\pi_\theta^{S}$; fixed teacher $\pi_{\bar{\theta}}^{T}$; composer $\Phi_\phi$ (init.\ from $\phi_0$); reward verifier $V\!:\boldsymbol{\tau}\mapsto[0,1]$; margin $m$; dual step $\eta_\beta$}
\Output{Trained student policy $\pi_\theta^{S}$}
$\beta\leftarrow 0$; \, cache $\boldsymbol{c}_{\phi_0}$ per task\;
\For{\rm{each training iteration}}{
    Sample tasks $\{x_i\}$ from $\mathcal{D}$\;
    \ForEach{\rm{task } $x_i$}{
        \tcc{\texttt{On-policy student rollout}}
        Sample trajectory $\boldsymbol{\tau}_i\sim \pi_\theta^{S}(\cdot\mid x_i)$;\quad $r_i\leftarrow V(\boldsymbol{\tau}_i)$;\quad $A_i\leftarrow 2r_i-1$\;
        \tcc{\texttt{Teacher evaluation with learned privileged context}}
        $\mathcal{E}_i \leftarrow \operatorname{Ret}(x_i)$;\;
        $\boldsymbol{c}_{\phi,i}\leftarrow \Phi_\phi(x_i,\mathcal{E}_i)$\;
        \ForEach{\rm{turn } $t$ and action prefix $a_{t,<n}$ in $\boldsymbol{\tau}_i$}{
            Compute $\boldsymbol{p}^{S}_{t,n}$ and $\boldsymbol{p}^{T}_{t,n}$ (\Cref{eq:multi-turn-teacher-student}); form top-$M$-plus-tail\;
            Accumulate $D_{\mathrm{KL}}(\tilde{\boldsymbol{p}}^S_{t,n}\Vert \tilde{\boldsymbol{p}}^T_{t,n})$\;
            $\delta_{t,n}\leftarrow \log\boldsymbol{p}^{T}_{t,n}(a_{t,n})-\operatorname{sg}[\log\boldsymbol{p}^{S}_{t,n}(a_{t,n})]$\;
        }
    }
    $\Delta\leftarrow$ weighted mean of $A_i\cdot\delta_{t,n}$ over masked tokens\;
    Update $(\theta,\phi)$ by minimizing \Cref{eq:lopd-objective}\;
    $\beta\leftarrow[\beta+\eta_\beta(m-\Delta)]_+$\;
}
\caption{Latent On-Policy Self-Distillation (\ourmethod).}
\label{alg:lopd}
\end{algorithm}

\vspace{-0.6em}
\paragraph{Algorithm Summary.}
\Cref{alg:lopd} summarizes the overall workflow of \ourmethod. The loop collects on-policy trajectories from the student, constructs learnable latent privileged context from retrieved experience, evaluates the same visited prefixes with the teacher, and jointly updates the student and composer while adjusting the dual variable to maintain the privilege margin. At inference, only the student policy $\pi_\theta^S$ is deployed and used.

\vspace{-0.4em}
\section{Experiments}
\vspace{-0.4em}
\subsection{Experiment Setup}
\vspace{-0.6em}
\paragraph{Training.}
We evaluate \ourmethod under two post-training settings: \ding{182} \textbf{agentic tool-use} and \ding{183} \textbf{coding}. For tool-use training, we use the EnvScaler-derived tool-interactive corpus~\citep{song2026envscalerscalingtoolinteractiveenvironments}, comprising $2{,}349$ tasks. For coding, we use the TACO subset of DeepCoder~\citep{togetherDeepCoderFully}, comprising $7$K verified Python problems (see \Cref{app:training-config} for training details). The used model backbones include \llmname{Qwen3-4B}, \llmname{Qwen3-8B}~\citep{yang2025qwen3}, \llmname{Olmo3-7B}~\citep{olmo2026olmo3}.

\vspace{-0.7em}
\paragraph{Evaluation.}
Our evaluation mirrors the two training regimes while keeping all test sets disjoint from training data. For tool-use, we report the EnvScaler task success metric on a held-out test split of $200$ tasks disjoint from the training pool, BFCL-v3 scores across base, missing-function, missing-parameter, and long-context subsets~\citep{patil2025bfcl}, and ACEBench multi-step and multi-turn scores~\citep{chen2025acebenchwinsmatchpoint}. For coding, we report pass@1 on LiveCodeBench v5/v6~\citep{jain2024livecodebench}, HumanEval+ and MBPP+~\citep{liu2023codegeneratedchatgptreally} (evaluation protocols in \Cref{app:eval-protocol}).

\vspace{-0.7em}
\paragraph{Baselines.}
We compare against the baselines reported in \Cref{tab:tool-use-results,tab:code-results}. \textbf{Vanilla} denotes the unadapted backbone. \textbf{GRPO} is the outcome-reward RL baseline~\citep{shao2024deepseekmath}. \textbf{SDFT}~\citep{shenfeld2026selfdistillationenablescontinuallearning} is a demonstration-conditioned distillation baseline. The OPSD-style baselines include \textbf{OPSD}~\citep{zhao2026selfdistilledreasoneronpolicyselfdistillation}, \textbf{SDPO}~\citep{hbotter2026reinforcementlearningselfdistillation}, and \textbf{Skill-SD}~\citep{wang2026skillsdskillconditionedselfdistillationmultiturn}, whose privileged contexts respectively instantiate answer/trace, feedback-conditioned self-teacher, and skill-conditioned supervision. Detailed baseline configurations are provided in \Cref{app:baseline-setup}.

\vspace{-0.7em}
\paragraph{Configurations.}
All trainable methods share the same backbone, training split and evaluation protocol within each setting. For \ourmethod, the teacher receives latent context composed from $J{=}3$ retrieved experiences, each compressed into $K{=}32$ latent tokens ($96$ total); the distillation loss uses reverse KL with SDPO-style top-$M$ logits plus a tail bucket, with $M{=}20$ by default. The privileged-margin threshold is $m{=}0.05$; the verification signal is $A(\boldsymbol{\tau})=2r(\boldsymbol{\tau})-1$ where $r$ is the environment reward. Tool-use rollouts are capped at $30$ environment steps. Coding distillation uses a $16{,}384$-token response budget. We keep decoding temperature/top-$p$ fixed across methods within each benchmark, each baseline retains its original KL direction and loss formulation (\Cref{app:baseline-setup}), and within the on-policy training loop the methods differ only in privileged context construction and distillation loss. Prompt templates are provided in \Cref{app:prompts}.

\begin{table}[!t]
\centering
\small
\setlength{\tabcolsep}{3pt}
\renewcommand{\arraystretch}{1.04}
\caption{
Tool-use results across EnvScaler, BFCL-v3, and ACEBench.
All results are reported on a $0$--$100$ scale; higher is better.
Green and light-green cells mark the best and second-best results.
}
    \vspace{-0.5em}
\label{tab:tool-use-results}
\resizebox{\textwidth}{!}{
\begin{tabular}{
ll@{\hspace{0.55em}}
c@{\hspace{0.65em}}
ccccc@{\hspace{0.65em}}
ccc
}
\toprule
\multirow{2}{*}{\textbf{LLM}} &
\multirow{2}{*}{\textbf{Method}} &
\multirow{2}{*}{\textbf{EnvScaler}} &
\multicolumn{5}{c}{\textbf{BFCL-v3}} &
\multicolumn{3}{c}{\textbf{ACEBench}} \\
\cmidrule(lr){4-8}
\cmidrule(lr){9-11}
& & &
\textbf{Base} & \textbf{Miss Func} & \textbf{Miss Param} & \textbf{Long Ctx} & \textbf{Avg}
& \textbf{M-Step} & \textbf{M-Turn} & \textbf{Avg} \\
\midrule
\multirow{7}{*}{\llmname{Qwen3-4B}}
& Vanilla
& 48.6
& 28.50 & \secondcell{20.50} & 20.50 & 22.00 & 22.88
& 48.3 & 52.2 & 50.6 \\
& GRPO
& \secondcell{61.8}
& \bestcell{33.00} & \secondcell{20.50} & \bestcell{24.50} & 23.00 & \secondcell{25.25}
& \bestcell{58.3} & 54.4 & \secondcell{56.0} \\
& Skill-SD
& 59.1
& \bestcell{33.00} & 19.00 & \secondcell{24.00} & 22.50 & 24.63
& \secondcell{56.6} & \secondcell{55.6} & \secondcell{56.0} \\
& OPSD
& 51.2
& 31.50 & 19.50 & 20.50 & \bestcell{29.00} & 25.13
& 41.6 & 53.3 & 48.6 \\
& SDPO
& 50.6
& 21.50 & 14.00 & 13.50 & 14.00 & 15.75
& 45.0 & 33.3 & 38.0 \\
& SDFT
& 50.3
& 31.50 & 17.00 & 17.00 & 21.50 & 21.75
& 53.3 & 47.8 & 50.0 \\
& \ourmethod
& \bestcell{63.7}
& \secondcell{32.50} & \bestcell{26.50} & \bestcell{24.50} & \secondcell{26.00} & \bestcell{27.38}
& \secondcell{56.6} & \bestcell{63.3} & \bestcell{60.6} \\
\midrule
\multirow{7}{*}{\llmname{Qwen3-8B}}
& Vanilla
& 49.2
& 34.00 & \secondcell{28.50} & 25.50 & \secondcell{25.50} & 28.38
& 61.6 & 50.0 & 54.6 \\
& GRPO
& 57.3
& \secondcell{36.00} & 28.00 & \secondcell{27.00} & 25.00 & \secondcell{29.00}
& \secondcell{65.0} & \secondcell{53.3} & \secondcell{58.0} \\
& Skill-SD
& \secondcell{60.2}
& 33.00 & 27.00 & 25.00 & 24.50 & 27.38
& 63.3 & 51.1 & 56.0 \\
& OPSD
& 52.0
& 30.50 & 25.00 & 23.50 & 24.00 & 25.75
& 60.0 & 47.8 & 52.7 \\
& SDPO
& 55.3
& 29.00 & 24.50 & 23.00 & 23.50 & 25.00
& 56.7 & 48.9 & 52.0 \\
& SDFT
& 56.2
& 31.50 & 26.00 & 24.50 & \secondcell{25.50} & 26.88
& \secondcell{65.0} & 47.8 & 54.7 \\
& \ourmethod
& \bestcell{66.4}
& \bestcell{37.00} & \bestcell{29.00} & \bestcell{27.50} & \bestcell{26.00} & \bestcell{29.88}
& \bestcell{73.3} & \bestcell{55.6} & \bestcell{62.7} \\
\bottomrule
\end{tabular}
}
\vspace{-0.4em}
\end{table}

\begin{table}[!t]
\centering
\small
\setlength{\tabcolsep}{9pt}
\renewcommand{\arraystretch}{1.04}
\caption{
Code results across LiveCodeBench and EvalPlus.
Each cell reports pass@1 (\%); higher is better.
Green and light-green cells mark the best and second-best within each backbone group.
}
    \vspace{-0.5em}
\label{tab:code-results}
\begin{tabular}{
ll@{\hspace{0.7em}}
ccc@{\hspace{1.0em}}
ccc
}
\toprule
\multirow{2}{*}{\textbf{LLM}} &
\multirow{2}{*}{\textbf{Method}} &
\multicolumn{3}{c}{\textbf{LiveCodeBench}} &
\multicolumn{3}{c}{\textbf{EvalPlus}} \\
\cmidrule(lr){3-5}
\cmidrule(lr){6-8}
& &
\textbf{v5} & \textbf{v6} & \textbf{Avg}
& \textbf{HumanEval+} & \textbf{MBPP+} & \textbf{Avg} \\
\midrule
\multirow{7}{*}{\llmname{Qwen3-4B}}
& Vanilla
& 47.67 & 41.22 & 45.61
& 85.37 & 75.93 & 78.79 \\
& GRPO
& \secondcell{50.18} & \secondcell{44.27} & \secondcell{48.29}
& 86.59 & 76.19 & 79.34 \\
& Skill-SD
& 49.82 & 41.98 & 47.32
& 85.98 & \secondcell{76.98} & 79.70 \\
& OPSD
& 42.65 & 35.11 & 40.24
& 85.36 & 76.72 & 79.33 \\
& SDPO
& 40.86 & 34.35 & 38.78
& 84.76 & 74.87 & 77.86 \\
& SDFT
& 48.75 & 43.51 & 47.07
& \secondcell{87.20} & \secondcell{76.98} & \secondcell{80.07} \\
& \ourmethod
& \bestcell{50.54} & \bestcell{45.04} & \bestcell{48.78}
& \bestcell{87.80} & \bestcell{78.57} & \bestcell{81.36} \\
\midrule
\multirow{7}{*}{\llmname{Olmo3-7B}}
& Vanilla
& 47.31 & 44.27 & 46.34
& 86.59 & 70.37 & 75.28 \\
& GRPO
& 49.46 & \bestcell{45.80} & \secondcell{48.29}
& \secondcell{89.02} & 72.75 & 77.67 \\
& Skill-SD
& 49.10 & 45.03 & 47.80
& 87.20 & 72.22 & 76.75 \\
& OPSD
& 44.09 & \secondcell{45.04} & 44.39
& 87.80 & 71.96 & 76.75 \\
& SDPO
& \secondcell{49.82} & 43.51 & 47.80
& 88.41 & 72.75 & 77.49 \\
& SDFT
& 47.67 & 44.27 & 46.58
& \secondcell{89.02} & \secondcell{73.02} & \secondcell{77.86} \\
& \ourmethod
& \bestcell{53.41} & \bestcell{45.80} & \bestcell{50.98}
& \bestcell{90.24} & \bestcell{73.28} & \bestcell{78.41} \\
\bottomrule
\end{tabular}
\vspace{-0.6em}
\end{table}

\vspace{-0.4em}
\subsection{Main Results}
\vspace{-0.6em}
We evaluate whether latent privileged context improves self-distillation learning across tool use and code generation domains in \Cref{tab:tool-use-results,tab:code-results}.

\vspace{-0.7em}
\paragraph{Performance.}
\ourmethod obtains the best aggregate result in all ten backbone--benchmark comparisons. On tool use with \llmname{Qwen3-4B}, it improves over the strongest competing method from $61.8$ to $63.7$ on EnvScaler, from $25.25$ to $27.38$ on BFCL-v3, and from $56.0$ to $60.6$ on ACEBench. The advantage becomes larger with \llmname{Qwen3-8B}: \ourmethod reaches $66.4/29.88/62.7$ on EnvScaler, BFCL-v3, and ACEBench, compared with the strongest baseline results of $60.2/29.00/58.0$, respectively. The same trend extends beyond interactive agents. With \llmname{Qwen3-4B}, \ourmethod improves the LiveCodeBench and EvalPlus aggregates to $48.78$ and $81.36$; with \llmname{Olmo3-7B}, it reaches $50.98$ and $78.41$, outperforming the strongest alternatives by $2.69$ and $0.55$ points. These results indicate that \ourmethod's improvement transfers across task formats and model families.

\vspace{-0.7em}
\paragraph{No Hand-Crafted Context Is Universally Optimal.}
A careful reader may notice that adding privileged information does not always improve upon the vanilla model in \Cref{tab:tool-use-results,tab:code-results}. For example, with \llmname{Qwen3-4B}, SDPO falls from $22.88$ to $15.75$ on BFCL-v3, from $50.6$ to $38.0$ on ACEBench, and from $45.61$ to $38.78$ on LiveCodeBench; OPSD likewise reduces the LiveCodeBench aggregate to $40.24$. More revealingly, the utility of the same context can reverse across settings: OPSD raises the BFCL-v3 long-context score from $22.00$ to $29.00$, yet remains below vanilla on the overall BFCL-v3 score with \llmname{Qwen3-8B} ($25.75$ vs. $28.38$) and on LiveCodeBench with both backbones ($40.24$ vs. $45.61$ and $44.39$ vs. $46.34$). This behavior is consistent with how these contexts are constructed. SDPO can condition only on successful siblings produced by the current rollout group, making its privileged signal dependent on the current policy's success coverage; OPSD instead injects a fixed oracle trace, which can be highly informative when its procedure matches the current task but need not align with the particular prefix or valid solution path visited by the student. A context format that helps in one regime can therefore become sparse, mismatched, or overly prescriptive in another. The issue is not simply whether privileged information is available, but whether its representation remains appropriate across tasks and learning states.

\vspace{-0.7em}
\paragraph{Learnable Context Matters.}
\ourmethod avoids committing the teacher to a fixed answer, demonstration, sibling rollout, or discrete skill. Instead, it learns a compact continuous representation directly from retrieved trajectories, allowing the privileged signal to adapt to the task and the student's visited prefixes. Unlike the reversals above, \ourmethod remains above the vanilla model in all ten aggregate backbone--benchmark settings, with gains ranging from $1.50$ points on BFCL-v3 with \llmname{Qwen3-8B} to $17.2$ points on EnvScaler with the same backbone. It also consistently improves over prescribed-context baselines: with \llmname{Qwen3-8B}, \ourmethod exceeds OPSD, SDFT, and Skill-SD by $14.4/10.2/6.2$ points on EnvScaler and $10.0/8.0/6.7$ points on ACEBench, respectively. Together, the results favor optimizing the representation of experience for supervision over searching for an increasingly elaborate hand-crafted artifact.

\begin{insightbox}{Takeaway}
Experience should not be treated as a designer-specified artifact, but as a learnable substrate for supervision. \ourmethod enables the agent to discover directly from trajectories what evidence to preserve and how to encode it, making experience learning end-to-end.
\end{insightbox}

\subsection{Framework Analysis}\label{sec:framework-analysis}
\vspace{-0.5em}

\vspace{-0.1em}
\begin{wrapfigure}{r}{0.4\textwidth}
    \vspace{-2.2em}
    \centering
    \includegraphics[width=\linewidth]{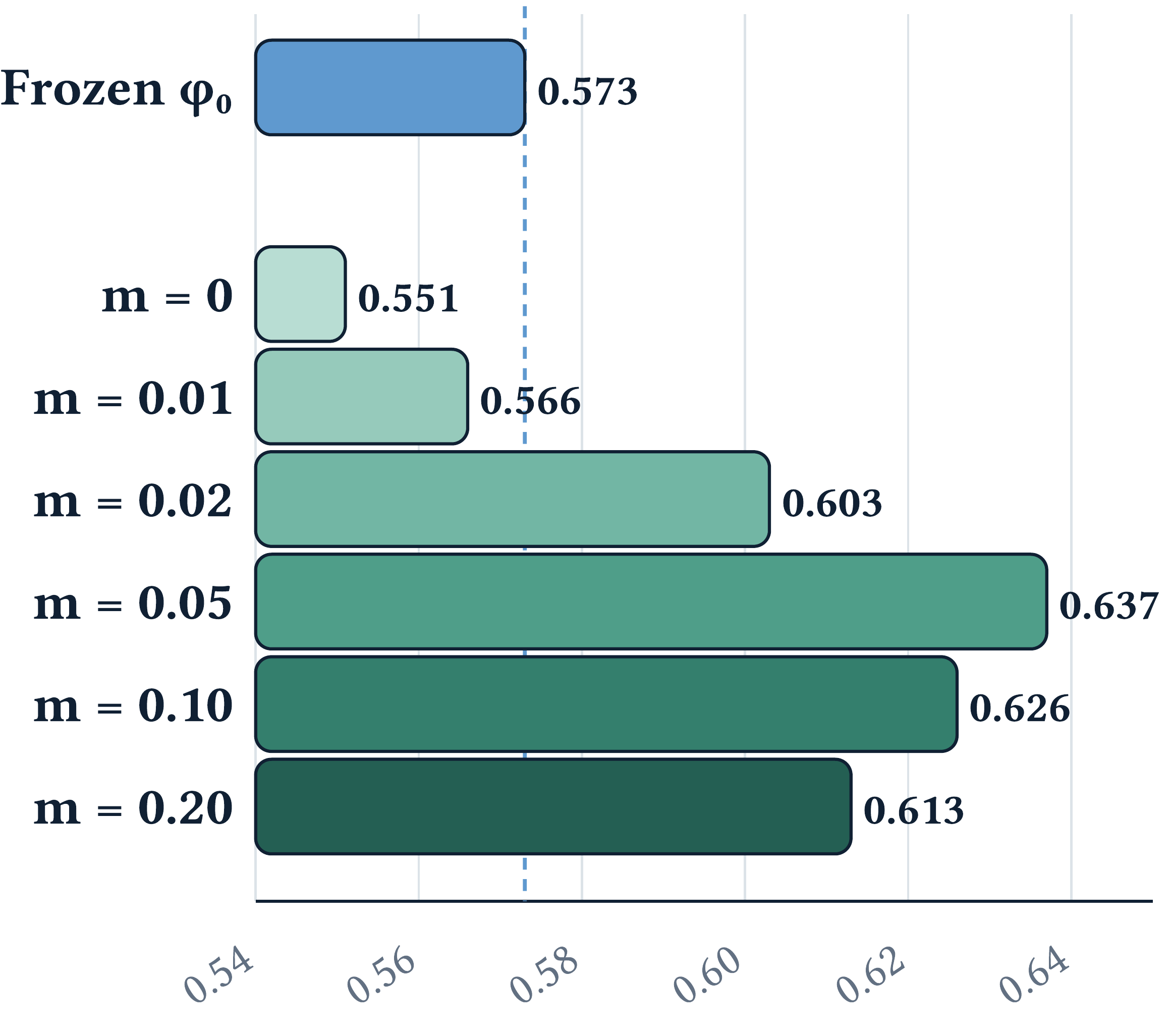}
    \vspace{-1.7em}
    \captionsetup{font=small,justification=raggedright,singlelinecheck=false}
    \caption{
    \textbf{Effect of joint optimization.}
    EnvScaler performance across margins.
    }
    \label{fig:composer-ablation}
    \vspace{-1.em}
\end{wrapfigure}
\tightparagraph{Effect of Joint Optimization.}
Does joint optimization actually produce a better privileged context, or merely introduce additional trainable parameters? This is the central ablation behind our claim. \Cref{fig:composer-ablation} compares a frozen composer with jointly optimized variants and evaluates each resulting student without retrieved experience or latent context, so the difference reflects what context learning transfers into the policy. Training with a frozen composer ($\phi_0$) achieves $0.573$. Without the margin constraint ($m{=}0$), the student drops to $0.551$, indicating that unconstrained distillation gradients degrade the latent context. Weak margins ($m{\le}0.01$) do not prevent this decline. With $m{\ge}0.02$, the student surpasses the frozen-composer baseline, reaching $0.637$ at $m{=}0.05$ and $0.626$ at $m{=}0.10$. The improvement suggests that the distillation process exposes the composer to a signal unavailable during initialization: what evidence the current student's trajectory distribution requires from the privileged context. The margin constraint is necessary to realize this benefit---without it, collapsing the teacher toward the student is a lower-resistance path than learning a more informative representation.

\begin{figure}[!t]
    \centering
    \includegraphics[width=0.98\textwidth]{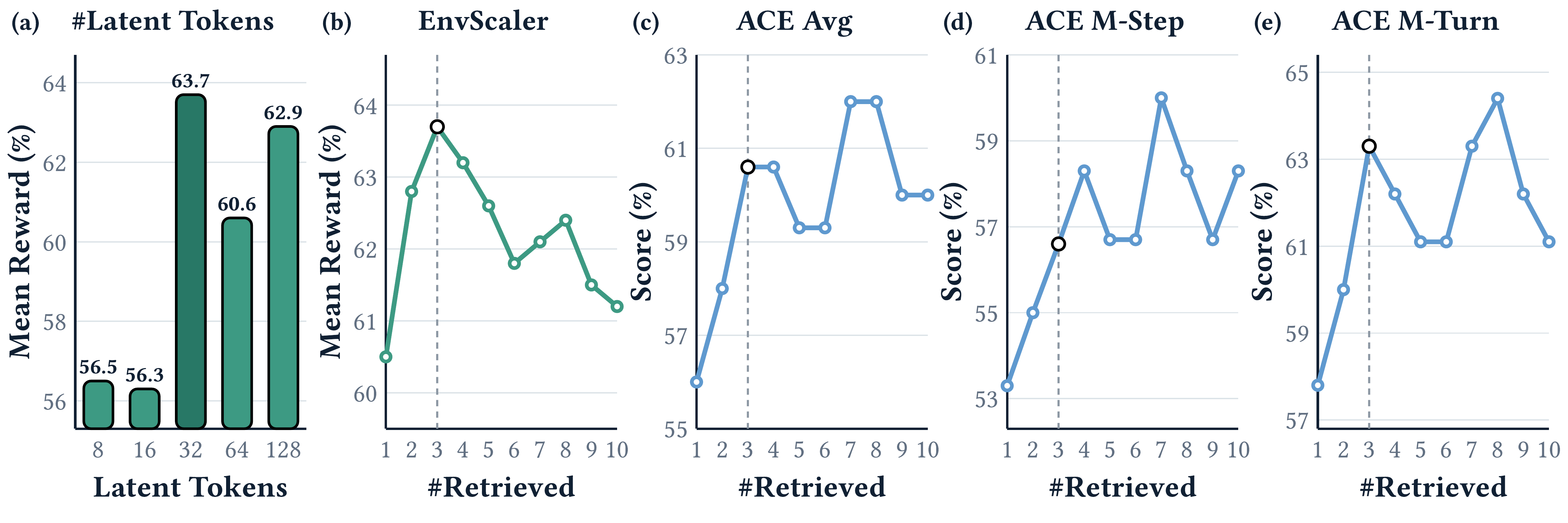}
    \vspace{-1.em}
    \caption{
    \textbf{Sensitivity analysis.}
    Performance of the resulting student without privileged context. (a) Latent-token capacity. (b) EnvScaler reward and (c--e) ACEBench aggregate, multi-step, and multi-turn scores as the training-time retrieval count varies. Dashed lines mark the default setting.
    }
    \label{fig:sensitivity-analysis}
    \vspace{-0.6em}
\end{figure}

\begingroup
\setlength{\intextsep}{16pt}
\makeatletter
\def\paragraph{\@startsection{paragraph}{4}{\z@}{1.5ex}{-1em}{\normalsize\bf}}
\makeatother
\begin{figure}[!t]
    \centering
    \begin{minipage}[t]{0.52\textwidth}
        \vspace{0pt}
        \centering
        \small
        \setlength{\tabcolsep}{9pt}
        \renewcommand{\arraystretch}{1.0}
        \resizebox{\linewidth}{!}{
        \begin{tabular}{lccc}
        \toprule
        \textbf{Metric} & \textbf{Vanilla} & \textbf{Base + Composer} & \textbf{\ourmethod} \\
        \midrule
        Reward & 0.486 & 0.631 & \textbf{0.637} \\
        Interaction steps & 11.12 & 16.31 & 17.04 \\
        Tool calls / step & 3.50 & 1.21 & \textbf{1.11} \\
        First-step length & 9,937 & 6,695 & \textbf{6,210} \\
        Reward / tool call & 0.038 & \textbf{0.053} & 0.050 \\
        Repeated tool calls & 8.89 & \textbf{4.49} & 5.25 \\
        \bottomrule
        \end{tabular}
        }
        \vspace{-0.4em}
        \captionof{table}{
        \textbf{Fine-grained behavior on the EnvScaler test set.}
        Base + Composer denotes the unadapted backbone conditioned on composed latent context; the student receives no privileged context.
        }
        \label{tab:behavior-analysis}
    \end{minipage}
    \hfill
    \begin{minipage}[t]{0.46\textwidth}
        \vspace{0pt}
        \centering
        \includegraphics[width=\linewidth]{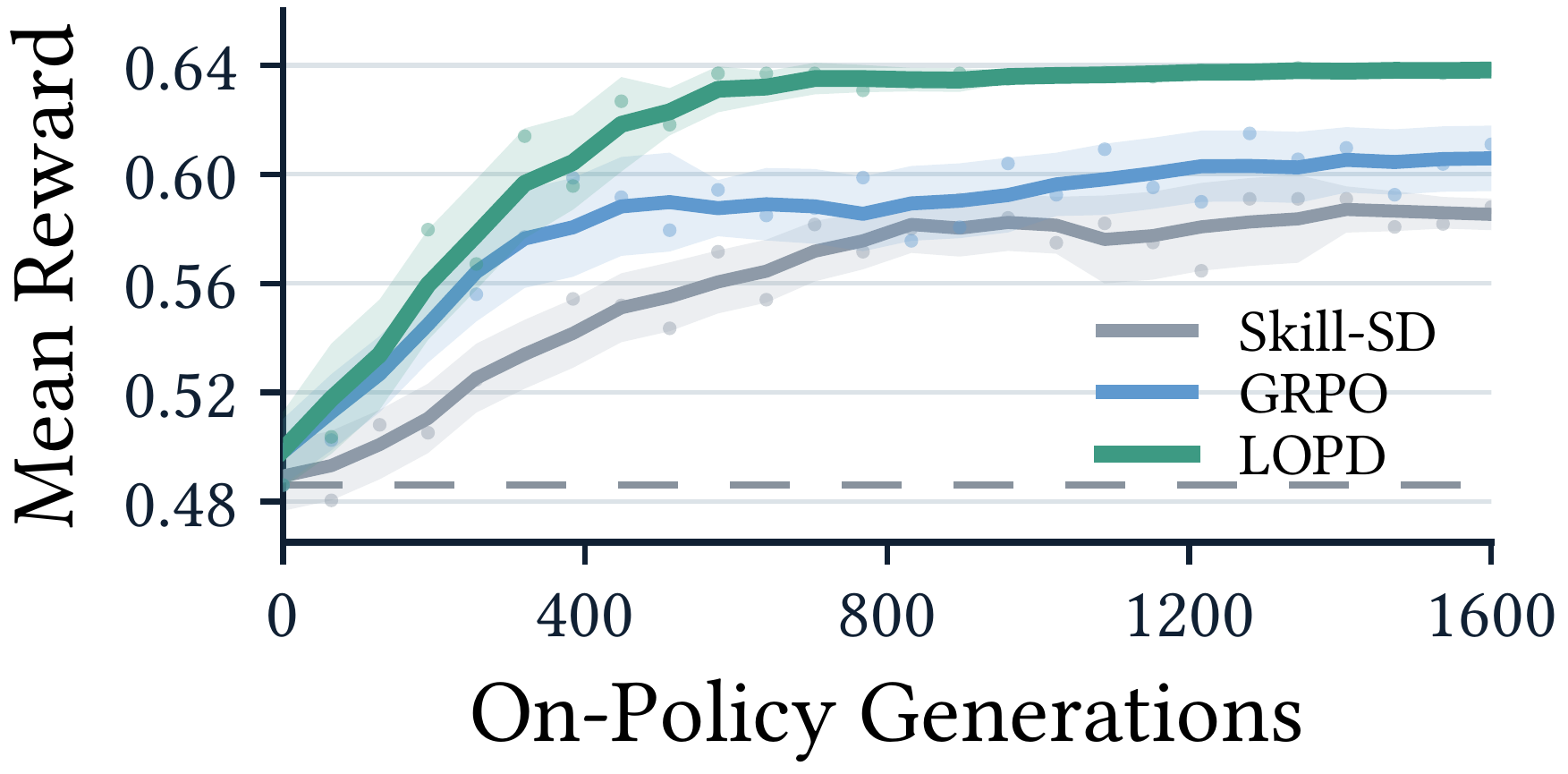}
        \vspace{-1.9em}
        \captionof{figure}{
        \textbf{EnvScaler training dynamics.}
        Mean reward over $1{,}600$ rollouts. 
        }
        \label{fig:envscaler-training-dynamics}
    \end{minipage}
    \vspace{-0.5em}
\end{figure}

\vspace{-0.4em}
\paragraph{Training Dynamics and Sample Efficiency.}
Beyond final performance, a practical question is whether \ourmethod reaches strong performance with fewer on-policy generations. To test this, \Cref{fig:envscaler-training-dynamics} tracks the strongest baselines over the same $1{,}600$-generation budget. \ourmethod exceeds $0.61$ mean reward after $320$ generations and reaches $0.637$ by generation $576$. Its reward then remains in a narrow $0.63$--$0.64$ range through generation $1{,}600$, showing that the early gain is sustained rather than caused by a shorter training horizon. GRPO and Skill-SD improve more gradually and finish at $0.611$ and $0.588$, respectively. The persistent early separation suggests that the latent teacher extracts a denser learning signal from each visited trajectory.

\vspace{-0.4em}
\paragraph{Sensitivity Analysis.}
How much latent capacity and retrieved experience does the learnable substrate actually need? We vary the number of latent tokens produced by the composer and the number of experiences retrieved during training, then evaluate the resulting student alone. \Cref{fig:sensitivity-analysis}(a) shows a capacity threshold in the latent bottleneck. EnvScaler reward remains near $0.56$ with $8$ or $16$ tokens per experience, rises sharply to $0.637$ with $32$, and then fluctuates without a consistent gain at $64$ and $128$. We therefore use $K{=}32$ tokens per experience as the smallest setting that escapes the low-capacity regime. Retrieval sensitivity is shown in \Cref{fig:sensitivity-analysis}(b--e), where $n_{\mathrm{ret}}$ varies from $1$ to $10$. EnvScaler reward improves from $0.605$ with one retrieval to $0.637$ with three, but additional retrievals yield no monotonic benefit. At the default $n_{\mathrm{ret}}{=}3$, ACEBench reaches $56.6$ on M-Step, $63.3$ on M-Turn, and $60.6$ overall, matching the main result in \Cref{tab:tool-use-results}. Larger retrieval counts can improve one ACEBench regime without producing the same trajectory in the other, and the aggregate remains on a broad plateau rather than varying monotonically. We therefore retain $n_{\mathrm{ret}}{=}3$ as the earliest setting that attains the strongest EnvScaler reward while already reaching competitive ACEBench performance.

\endgroup

\vspace{-0.4em}
\paragraph{Behavioral Internalization.}
Finally, higher reward alone does not reveal what the student has internalized. We therefore ask whether \ourmethod changes how the student interacts with the environment. \Cref{tab:behavior-analysis} shows that the distilled student inherits the interaction pattern induced by latent context. Relative to the vanilla model, \ourmethod uses more environment steps ($17.04$ vs. $11.12$) but makes far fewer tool calls per step ($1.11$ vs. $3.50$), indicating a shift from issuing many speculative calls at once to executing a more sequential plan. Its first-step response is $37.5\%$ shorter, repeated calls fall from $8.89$ to $5.25$, and reward per tool call rises from $0.038$ to $0.050$. The latent-context-conditioned base model and the final student exhibit the same qualitative pattern, providing behavioral evidence that \ourmethod internalizes the teacher's procedural guidance rather than merely fitting the aggregate reward. Per-experiment configurations are detailed in \Cref{app:ablation-setup}.

\begin{insightbox}{Mechanistic Takeaway}
Outcome reward aligns the teacher's aggregate advantage with trajectories that achieve better outcomes, while on-policy distillation converts the resulting context-conditioned guidance into dense supervision at states the student actually visits. The guidance is internalized when it persists after privileged context is removed and reappears as a stable change in the student's behavior.
\end{insightbox}

\vspace{-0.1em}
\subsection{Case Study}\label{sec:case-study}
\begin{figure}[!h]
    \centering
    \includegraphics[width=\textwidth]{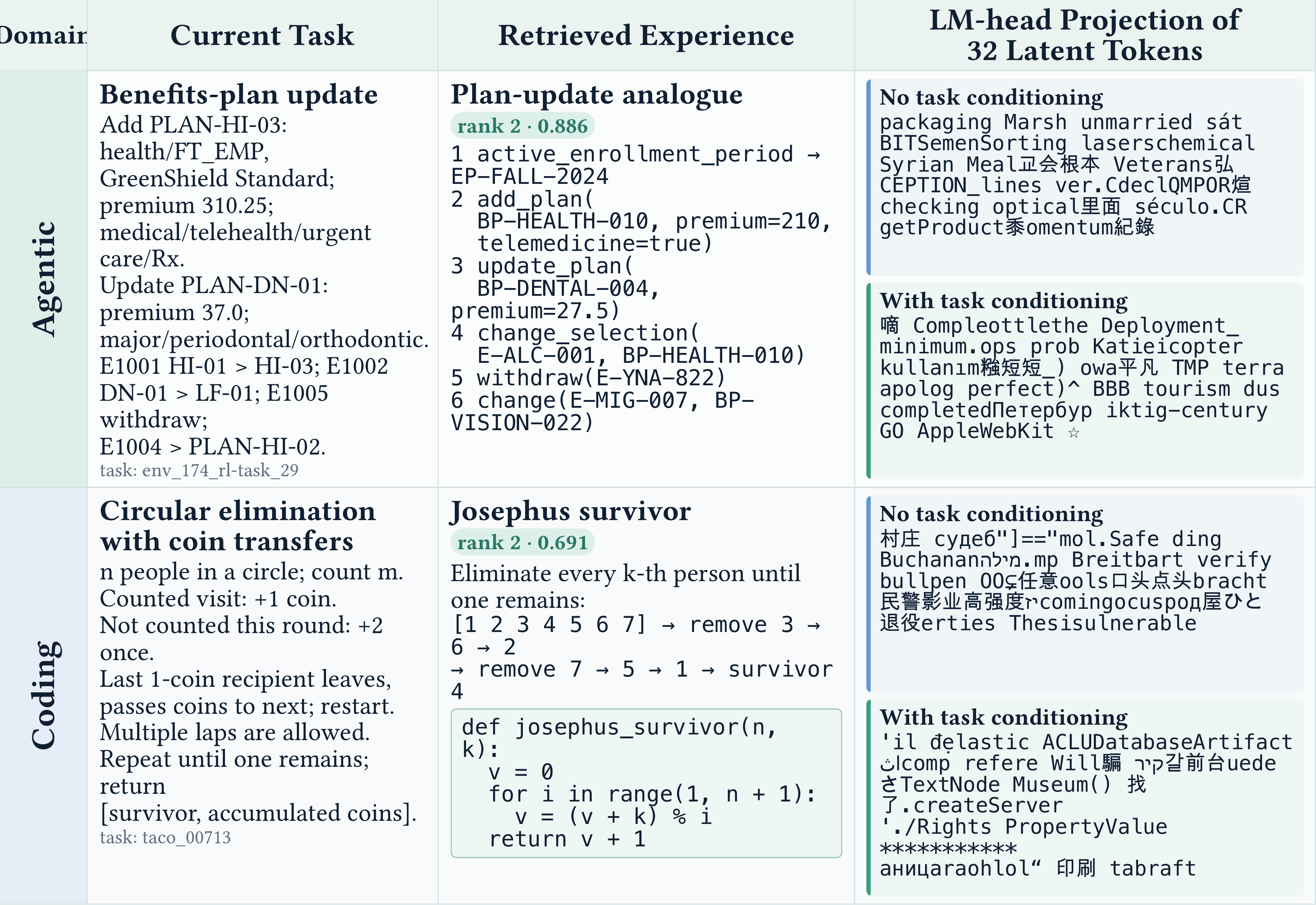}
    \vspace{-0.8em}
    \caption{
    \textbf{Representative latent decoding examples.}
    Each row pairs a current task with a structurally related rank-2 retrieval and excerpts from LM-head projections of $32$ latent tokens, with and without task conditioning. The projections remain fragmented and do not reproduce the retrieved procedure verbatim.
    }
    \label{fig:latent-decode-case-study}
\end{figure}

\vspace{-0.4em}
Having established that the latent context improves learning, a natural question is what these continuous tokens actually encode. To obtain a qualitative view, we apply the frozen language-model head to each of the $32$ latent tokens and inspect ten tool-use and ten coding examples, with and without task conditioning. \Cref{fig:latent-decode-case-study} shows two representative cases. In the agentic example, the current task and retrieved trajectory share the same add--update--change--withdraw operation schema despite using different entities. In the coding example, the retrieved Josephus recurrence matches the circular-elimination structure of the target problem. Nevertheless, both projections remain fragmented mixtures of multilingual and code-like tokens, and task conditioning changes the surface projection without yielding a readable procedure or copying the retrieved solution. This is consistent with a distributed latent representation, although direct decodability alone does not establish which information the teacher functionally uses.

\section{Conclusion}
\vspace{-0.4em}
The question behind this work is not which new artifact should be appended to an OPSD teacher, but whether the teacher's privileged context can itself be learned from experience. \ourmethod answers this question by transforming retrieved raw trajectories into differentiable latent privileged context, using the resulting self-teacher to supervise the student's own visited prefixes, and constraining the learned context to preserve a verifiable teacher advantage. Across agentic tool use and code generation, this formulation achieves the best aggregate result in all ten backbone--benchmark comparisons and surpasses GRPO and Skill-SD with less than $30\%$ of their rollout budget. The analyses sharpen this interpretation: retrieval alone is insufficient, unconstrained joint optimization can collapse the teacher toward the student, and the privileged margin turns context learning into productive supervision. The induced behavioral shift remains in the trained student, which acts without privileged context. More broadly, scalable self-evolution should not depend on a succession of increasingly elaborate, human-authored experience formats. Raw trajectories provide a minimal substrate, and richer repositories or retrievers may broaden what is available, but learning should decide what becomes useful guidance. In this sense, \ourmethod is less another privileged-context recipe than evidence for a different design principle: experience representations should be optimized end-to-end for the policies they are meant to improve.

\bibliography{iclr2026_conference}
\bibliographystyle{iclr2026_conference}

\appendix

\section{Implementation Details}

\subsection{Composer Architecture}\label[appendix]{app:composer-arch}

The latent composer $\Phi_\phi$ consists of an encoder $\operatorname{Enc}_\psi$ and a compressor $\operatorname{Comp}_\chi$, with $\phi\coloneqq(\psi,\chi)$. The encoder uses the frozen backbone augmented with a trainable LoRA adapter (rank $8$, $\alpha{=}16$, dropout $0$, applied to all seven linear projections per transformer block: $\mathbf{W}_q$, $\mathbf{W}_k$, $\mathbf{W}_v$, $\mathbf{W}_o$, gate, up, and down projections). During encoding, the task description is prepended to each retrieved experience to enable task-conditional compression, so that the QFormer's cross-attention can condition on the current task when compressing the experience (see \Cref{app:prompts} for the exact format).

The compressor is a QFormer-style perceiver where learned queries cross-attend to the encoder's hidden states. It has $8$ cross-attention layers with shared parameters (i.e., all layers reuse the same weights), attention heads and hidden dimension inherited from the backbone model, and a feed-forward multiplier of $4$. Learned queries $\mathbf{Q}_\chi \in \mathbb{R}^{K \times d}$ are initialized from $\mathcal{N}(0, 1/\sqrt{d})$. The compressor operates independently on each retrieved experience, producing $K$ latent tokens per item; the outputs are concatenated to form the full latent context $\boldsymbol{c}_\phi$. Based on the sensitivity analysis in \Cref{sec:framework-analysis}, we set $K{=}32$ tokens per retrieved experience for the final \ourmethod training runs. The encoder LoRA and QFormer are updated during both cold-start and joint optimization; the backbone weights remain frozen throughout.

\subsection{Cold-Start Initialization}\label[appendix]{app:cold-start}

The composer is cold-started by supervised finetuning on retrieved-experience--trajectory pairs with frozen backbone. The training objective is next-token NLL on assistant tokens with latent-augmented input:
\begin{equation}
\mathcal{L}_{\mathrm{init}}(\phi)=-\mathbb{E}_{(x,\boldsymbol{y}^\star)\sim\mathcal{D}_{\mathrm{exp}},\;\mathcal{E}=\operatorname{Ret}(x)}\!\left[\log\pi_{\bar{\theta}}\!\left(\boldsymbol{y}^\star\mid\boldsymbol{s},\boldsymbol{c}_\phi(x,\mathcal{E})\right)\right],
\label{eq:cold-start}
\end{equation}
where $\boldsymbol{y}^\star$ is a successful trajectory and $\bar{\theta}$ denotes the frozen backbone. Concretely, the composer first produces latent tokens $\boldsymbol{c}_\phi$ from retrieved experiences and inserts them into the input embedding sequence at designated placeholder positions. The frozen backbone then performs a standard forward pass over this latent-augmented input. Although the backbone parameters receive no gradient updates, the loss gradient flows backward through the frozen layers' activations to the latent token positions, and from there to the QFormer and LoRA parameters---the same mechanism as prefix-tuning.

The training data is synthesized from the base model's own rollouts---no external expert or stronger model is required---and filtered by task success. The data volume and number of training steps are adjusted per domain. Training uses AdamW with learning rate $10^{-5}$, batch size $8$, gradient clipping $3.0$, task-conditional compression, and $n_\mathrm{ret}{=}3$ retrieved experiences per task. The LoRA adapter and QFormer are updated during cold-start and remain trainable during subsequent joint optimization; the backbone remains frozen.

\subsection{Retrieval Configuration}\label[appendix]{app:retrieval}

Let $\mathcal{B}=\{(x_i, \boldsymbol{\tau}_i)\}_{i=1}^{|\mathcal{B}|}$ denote the experience bank, where each entry pairs a task description $x_i$ with a successful trajectory $\boldsymbol{\tau}_i$. A dense encoder $f_\mathrm{ret}$ maps text to a unit-norm embedding. We adopt asymmetric encoding: document-side embeddings encode the concatenation of task description and trajectory, while query-side embeddings encode only the task description with an instruction-aware query prompt:
\begin{equation}
\boldsymbol{z}_i^{\mathrm{doc}} = \frac{f_\mathrm{ret}([x_i; \boldsymbol{\tau}_i])}{\|f_\mathrm{ret}([x_i; \boldsymbol{\tau}_i])\|_2}, \qquad
\boldsymbol{z}_q = \frac{f_\mathrm{ret}^{\mathrm{query}}(x)}{\|f_\mathrm{ret}^{\mathrm{query}}(x)\|_2}, \qquad
\boldsymbol{z}_i^{\mathrm{doc}},\, \boldsymbol{z}_q \in \mathbb{R}^{d_\mathrm{ret}}.
\end{equation}
Given a query task $x$, retrieval returns the $n_\mathrm{ret}$ entries with the highest cosine similarity:
\begin{equation}
\operatorname{Ret}(x) = \operatorname{top\text{-}n_\mathrm{ret}} \left\{ (x_i, \boldsymbol{\tau}_i) \in \mathcal{B} \;\middle|\; {\boldsymbol{z}_q}^\top \boldsymbol{z}_i^{\mathrm{doc}} \right\}.
\end{equation}
We use \llmname{Qwen3-Embedding-8B} as $f_\mathrm{ret}$, producing $d_\mathrm{ret}{=}4{,}096$-dimensional embeddings. The index is implemented as a FAISS \texttt{IndexFlatIP} for exact inner-product search over the precomputed document embeddings. The bank $\mathcal{B}$ is constructed offline exclusively from successful rollouts generated on the training split: agentic trajectories must exceed a task-reward threshold, while coding trajectories must pass all test cases. No evaluation task, trajectory, or outcome is ever inserted into the bank, and the bank is frozen before evaluation. Consequently, \ourmethod receives neither evaluation-set information nor an additional task corpus; it uses only experience produced from the same training-task split available to the baselines. Trajectories are stored in an observation-lite format that omits verbose environment observations to reduce storage and encoding length. All query embeddings $\boldsymbol{z}_q$ are precomputed and cached per task, so the retriever does not need to be loaded during training or evaluation.

Each bank entry stores the task description followed by a linearized action--result trace in observation-lite format, as illustrated below:

\begin{tcolorbox}[colback=SeaGreen!6, colframe=SeaGreen!50!black, fontupper=\ttfamily\scriptsize, boxrule=0.5pt, left=4pt, right=4pt, top=3pt, bottom=3pt, title={\small\bfseries Example experience-bank entry (agentic, truncated)}]
\textbf{Task:}\\
Acting on behalf of Alice Chan, update her health policy XJ-439220: add Psychiatric Care coverage (\$3500 limit, \$300 deductible), increase ER coverage limit to \$7000, ...\\[0.3em]
\textbf{Example steps} (from a different task, NOT yet executed):\\
1. get\_policy\_by\_policy\_number("XJ-439220") -> policy\_id=POL-001\\
2. get\_exclusions\_for\_policy("POL-001") -> exclusion\_id=EXCL-001, ...\\
3. remove\_exclusion\_from\_policy("POL-001", "EXCL-002") -> Success\\
4. add\_coverage\_item\_to\_policy("POL-001", ...) -> Coverage item added\\
5. update\_coverage\_limit\_or\_deductible("COV-002", limit=7000) -> Updated\\
...
\end{tcolorbox}

\subsection{Latent Injection Mechanism}\label[appendix]{app:latent-injection}

The latent tokens produced by the composer must be injected into the backbone's input in a way compatible with the inference engine's pipeline. Unlike prior latent-state injection work that typically relies on HuggingFace Transformers or TRL for direct \texttt{inputs\_embeds} manipulation, our implementation operates on top of production inference engines (SGLang and vLLM).

\baselineparagraph{Placeholder strategy.}
A text-level sentinel string \texttt{<|LATENT\_PH|>} is inserted into the first user message of the chat template, sandwiched between natural-language framing text (e.g., ``\textit{The following is a reference example\ldots}'' before and ``\textit{Now complete your task\ldots}'' after). The full prompt is then rendered via the tokenizer's chat template and split on the sentinel. Each half is tokenized independently, and $J \cdot K$ dummy token IDs (using the tokenizer's pad token) are inserted between them:
\begin{equation}
\mathrm{input\_ids} = [\,\underbrace{t_1, \ldots, t_a}_{\text{before}},\; \underbrace{t_{\mathrm{pad}}, \ldots, t_{\mathrm{pad}}}_{J \cdot K \text{ placeholders}},\; \underbrace{t_{a+1}, \ldots, t_L}_{\text{after}}\,].
\end{equation}
This produces a contiguous span of placeholder positions at known absolute indices, without any tokenizer modification.

\baselineparagraph{Embedding replacement.}
The engine embeds the full \texttt{input\_ids} as usual, producing $\mathbf{X} \in \mathbb{R}^{L \times d}$. Before the first transformer layer, the placeholder embeddings are replaced with the latent tokens from the composer:
\begin{equation}
\mathbf{X}[p_k] \leftarrow \langle e_k \rangle, \quad k = 1, \ldots, J \cdot K.
\end{equation}
The replacement uses \texttt{torch.cat} over slices (rather than in-place assignment) so that autograd can track gradients through $\langle e_k \rangle$ back to the composer parameters during training. The modified embedding tensor is then passed through the backbone's transformer layers without any architectural change; the operation is transparent to the engine's KV-cache, attention mask, and batching logic.

\baselineparagraph{Engine-specific implementations.}
SGLang provides a \texttt{positional\_embed\_overrides} interface that accepts embeddings at specified absolute positions and performs the replacement in-GPU.
vLLM provides a \texttt{prompt\_embeds} interface: the full sequence is first embedded on CPU via a frozen copy of the embedding table, the latent positions are overwritten with the composer's output, and the complete embedding tensor is submitted to the engine.
In both cases, the backbone model is unmodified.

\baselineparagraph{Equivalence verification.}
To confirm that the placeholder-based injection introduces no numerical artifacts, we replaced the latent span with known token embeddings from the vocabulary and compared the token-ID input path against the embedding-tensor input path. The two paths produce identical greedy-decoded sequences with negligible log-probability differences, verifying that latent injection is numerically equivalent to standard token-ID forwarding.

\baselineparagraph{Encoder LoRA isolation.}
The encoder $\operatorname{Enc}_\psi$ uses a LoRA adapter to specialize hidden-state extraction for experience compression. This adapter is explicitly \emph{disabled} during the main forward pass (both at inference and during teacher evaluation in training), so that the model behaves as a pure base actor conditioned on the injected latent context. This dual-use pattern---LoRA active for encoding, inactive for generation---avoids interference between the two roles.

\subsection{Prompt and Interaction Format}\label[appendix]{app:prompts}

\paragraph{Agentic system prompt.}
For tool-use tasks (EnvScaler, ACEBench), the agent receives a system prompt instructing it to complete tasks via step-by-step tool invocation:

\begin{tcolorbox}[colback=SeaGreen!6, colframe=SeaGreen!50!black, fontupper=\ttfamily\scriptsize, boxrule=0.5pt, left=4pt, right=4pt, top=3pt, bottom=3pt, title={\small\bfseries Agentic system prompt (EnvScaler / ACEBench)}]
You are a helpful assistant. When given a specific task, your goal is to complete it in an interactive environment by making step-by-step use of available tools.\\
- Before completing the task, at each step, select a tool from the tool list and fill in all required parameters, making sure that the values are valid. Avoid making parallel tool calls in one step.\\
- When you believe the task has been completed, respond only with `Task Completed' to end the trajectory, without adding any other content or making any tool calls.\\
- It is recommended to first call query tools to gather sufficient information, then use modification tools to complete the task. Adjust actions promptly based on the feedback from the environment.
\end{tcolorbox}

\paragraph{Coding system prompt.}
For TACO and LiveCodeBench, the model receives a minimal system instruction:

\begin{tcolorbox}[colback=SeaGreen!6, colframe=SeaGreen!50!black, fontupper=\ttfamily\scriptsize, boxrule=0.5pt, left=4pt, right=4pt, top=3pt, bottom=3pt, title={\small\bfseries Coding system prompt (TACO / LiveCodeBench)}]
You are an expert Python programmer. You will be given a question (problem specification) and will generate a correct Python program that matches the specification and passes all tests.
\end{tcolorbox}

\noindent HumanEval+ and MBPP+ use no system prompt, following the official EvalPlus evaluation protocol. All coding benchmarks use Python as the target language.

\paragraph{Latent-context framing text.}
The latent privileged context is injected into the first user message, wrapped by domain-specific framing text. The framing instructs the model to treat the injected content as a reference from a different task, not as instructions to follow verbatim.

\begin{tcolorbox}[colback=SeaGreen!6, colframe=SeaGreen!50!black, fontupper=\ttfamily\scriptsize, boxrule=0.5pt, left=4pt, right=4pt, top=3pt, bottom=3pt, title={\small\bfseries Agentic privileged-context framing (EnvScaler)}]
\textbf{Before latent context:}\\
The following is a REFERENCE EXAMPLE of how a DIFFERENT, already-solved task was handled, shown only to illustrate the general approach. It was performed in a SEPARATE session -- in YOUR task below, NOTHING has been done yet and the environment is in its initial state. You must perform every step yourself by calling the tools and reading their actual results; do NOT assume any step is already complete.\\[0.3em]
\textbf{After latent context:}\\
--- end of reference example ---\\
Now complete YOUR task below, starting from scratch (the environment is untouched):
\end{tcolorbox}

\begin{tcolorbox}[colback=SeaGreen!6, colframe=SeaGreen!50!black, fontupper=\ttfamily\scriptsize, boxrule=0.5pt, left=4pt, right=4pt, top=3pt, bottom=3pt, title={\small\bfseries Coding privileged-context framing (TACO)}]
\textbf{Before:}\\
The following is a reference from a similar programming problem. Study the approach and algorithm, then solve YOUR problem below.\\[0.3em]
\textbf{After:}\\
--- end of reference ---
\end{tcolorbox}

The latent tokens produced by the composer replace the framing's interior (as described in \Cref{app:latent-injection}).

\paragraph{Encoder input format.}
When the encoder $\operatorname{Enc}_\psi$ processes a retrieved experience for compression, the input is formatted as ``\texttt{Task to solve:}\textbackslash n$x$\texttt{\textbackslash n\textbackslash nReference past trajectory:}\textbackslash n$m_j$'', where $x$ is the current task description and $m_j$ is the retrieved trajectory. This task-conditional formatting allows the QFormer to attend to task-relevant features when producing the latent tokens.

\section{Experiment Details}

\subsection{Baseline Setup}\label[appendix]{app:baseline-setup}

All trainable methods share the same base model, training task distribution, on-policy rollout budget ($32$ rollouts per step), optimizer (AdamW, learning rate $10^{-5}$, gradient clipping $1.0$), and evaluation protocol. Each distillation-based method uses the KL direction prescribed by its original paper: OPSD and SDFT use forward KL over the full vocabulary, SDPO uses reverse KL with top-$K$ truncation and a tail bucket, and Skill-SD uses sampled-token reverse KL with importance weighting. SDFT and OPSD require ground-truth demonstrations as privileged context for the teacher. These are drawn from a shared pool of verified-successful trajectories: for agentic tasks, trajectories must exceed a reward threshold; for coding tasks, solutions must pass all test cases. The pool is first populated from the base model's own rollouts; when its coverage is insufficient, we synthesize additional ground-truth trajectories by querying stronger models (DeepSeek-V4-Pro and Qwen3.7-Max in our experiments). Below we describe the method-specific configurations.

\methodparagraph{Vanilla.}
The unadapted backbone model, evaluated without any post-training.

\methodparagraph{GRPO.}
Standard group relative policy optimization~\citep{shao2024deepseekmath}. Each step samples $G{=}4$ rollouts per task with stochastic decoding; group-normalized advantages weight a PPO-clip objective with $\epsilon_\mathrm{lo}{=}\epsilon_\mathrm{hi}{=}0.2$. No teacher or privileged context is used; the training signal comes entirely from environment rewards.

\methodparagraph{SDFT.}
Demonstration-conditioned distillation~\citep{shenfeld2026selfdistillationenablescontinuallearning}. The teacher receives oracle trajectories from the shared pool as textual in-context demonstrations appended to the first user message. The teacher is an EMA shadow of the student ($\alpha_\mathrm{EMA}{=}0.01$, synced every step). The distillation loss is forward KL over the full vocabulary, following the official implementation.

\begin{tcolorbox}[before skip=4pt, after skip=1pt, colback=SeaGreen!6, colframe=SeaGreen!50!black, fontupper=\ttfamily\scriptsize, boxrule=0.5pt, left=4pt, right=4pt, top=3pt, bottom=3pt, title={\small\bfseries SDFT demonstration injection format}]
\textbf{Appended to user message:}\\
This is an example for a response to the question:\\
\{demo\_text\}\\[0.3em]
Now answer with a response of your own, including the thinking process.
\end{tcolorbox}

\methodparagraph{OPSD.}
On-policy self-distillation~\citep{zhao2026selfdistilledreasoneronpolicyselfdistillation}. The teacher receives oracle trajectories from the shared pool directly as privileged context in the system message. The teacher is frozen throughout training. The distillation loss is forward KL over the full vocabulary, as prescribed by the original paper.

\begin{tcolorbox}[before skip=4pt, after skip=1pt, colback=SeaGreen!6, colframe=SeaGreen!50!black, fontupper=\ttfamily\scriptsize, boxrule=0.5pt, left=4pt, right=4pt, top=3pt, bottom=3pt, title={\small\bfseries OPSD GT injection format}]
\textbf{Appended to system message:}\\
The following is a reference plan for the upcoming task. Use it only as guidance for your own reasoning; you must still interact with the environment to complete the task.\\
\{gt\_text\}\\[0.3em]
After reading the reference solution above, make sure you truly understand the reasoning behind each step. Now, using your own words and independent reasoning, derive the same final answer.
\end{tcolorbox}

\methodparagraph{SDPO.}
Self-distillation policy optimization~\citep{hbotter2026reinforcementlearningselfdistillation}. The teacher is an EMA shadow of the student (update rate $0.05$), conditioned on successful sibling trajectories sampled within the same training step ($G{=}4$ rollouts per task). Unlike SDFT and OPSD, SDPO does not use an external trajectory pool; it filters from its own rollouts using the same domain-specific success criterion (reward threshold for agentic tasks, full test-case pass for coding tasks).

\methodparagraph{Skill-SD.}
Skill-conditioned self-distillation~\citep{wang2026skillsdskillconditionedselfdistillationmultiturn}. Each task selects one skill from a skill bank via UCB1 ($c{=}\sqrt{2}$). Skills are distilled from rollout trajectories into structured summaries and injected into the teacher's first user message:

\begin{tcolorbox}[before skip=4pt, after skip=6pt, colback=SeaGreen!6, colframe=SeaGreen!50!black, fontupper=\ttfamily\scriptsize, boxrule=0.5pt, left=4pt, right=4pt, top=3pt, bottom=3pt, title={\small\bfseries Skill-SD skill injection format}]
\textbf{Appended to user message:}\\
Here is a skill summary from a similar past task that may help guide your approach:\\[0.3em]
**What works:** \{success\_analysis\}\\
**What to avoid:** \{mistake\_analysis\}\\
**Suggested workflow:** \{golden\_workflow\}\\[0.3em]
Now solve the current task using your own reasoning.
\end{tcolorbox}

\noindent The loss combines PPO-clip ($\epsilon_\mathrm{lo}{=}0.2$, $\epsilon_\mathrm{hi}{=}0.28$) with importance-weighted sampled-token reverse KL ($\lambda_\mathrm{sdl}{=}0.001$), following the original paper.

\subsection{Training Data and Domain Setup}\label[appendix]{app:training-config}

All methods are trained exclusively on the two datasets listed in \Cref{tab:training-config}; no additional task dataset is used. In particular, the \ourmethod experience bank contains only rollouts from the corresponding training split and excludes every evaluation task and trajectory. The agentic training corpus (EnvScaler) covers diverse tool-interactive scenarios including insurance, logistics, e-commerce, and healthcare, with each task requiring multi-turn API interactions to complete. The coding training corpus (TACO subset of DeepCoder) consists of competitive programming problems with verified test cases, spanning algorithmic topics such as dynamic programming, graph traversal, and string manipulation. The environment reward differs by domain: EnvScaler returns a continuous reward in $[0,1]$ reflecting the fraction of subtasks completed, while TACO returns a binary reward ($1$ if all test cases pass, $0$ otherwise).

\begin{table}[H]
\centering
\small
\setlength{\tabcolsep}{10pt}
\renewcommand{\arraystretch}{1.08}
\begin{tabular}{lcc}
\toprule
\textbf{Parameter} & \textbf{Agentic} & \textbf{Coding} \\
\midrule
Training data & EnvScaler & TACO (DeepCoder) \\
Training tasks & $2{,}349$ & ${\sim}7{,}000$ \\
Total generation budget (tokens) & $40{,}960$ & $16{,}384$ \\
Max environment steps & $30$ & --- (single-turn) \\
Decoding temperature & \multicolumn{2}{c}{$0.7$} \\
top-$p$ & \multicolumn{2}{c}{$0.95$} \\
top-$k$ (sampling) & \multicolumn{2}{c}{$20$} \\
\midrule
\multicolumn{3}{l}{\textit{\ourmethod constraint parameters}} \\
Composer learning rate ($\mathrm{lr}_\phi$) & \multicolumn{2}{c}{$10^{-5}$} \\
Anchor weight ($\lambda$) & \multicolumn{2}{c}{$0.2$} \\
Dual step size ($\eta_\beta$) & \multicolumn{2}{c}{$0.5$} \\
\bottomrule
\end{tabular}
\caption{Domain-specific training configurations.}
\label{tab:training-config}
\end{table}

\subsection{Evaluation Protocols}\label[appendix]{app:eval-protocol}

\Cref{tab:eval-protocol} summarizes the inference configuration for each benchmark. All benchmarks use thinking mode enabled.

\begin{table}[H]
\centering
\small
\setlength{\tabcolsep}{4pt}
\renewcommand{\arraystretch}{1.08}
\begin{tabular}{lcccccc}
\toprule
\textbf{Benchmark} & \textbf{Tasks} & \textbf{Temp} & \textbf{top-$p$} & \textbf{top-$k$} & \textbf{Budget (tokens)} & \textbf{Max steps} \\
\midrule
EnvScaler & $200$ & $0.7$ & $0.95$ & $20$ & $40{,}960$ & $30$ \\
BFCL-v3 & $4{\times}200$ & $0.7$ & $0.95$ & $20$ & $40{,}960$ & $30$ \\
ACEBench & $50$ & $0.7$ & $0.95$ & $20$ & $40{,}960$ & $20$ \\
LiveCodeBench & $279{+}131$ & $0.7$ & $0.95$ & $20$ & $16{,}384$ & --- \\
HumanEval+ & $164$ & $0.7$ & $0.95$ & $20$ & $16{,}384$ & --- \\
MBPP+ & $378$ & $0.7$ & $0.95$ & $20$ & $16{,}384$ & --- \\
\bottomrule
\end{tabular}
\caption{Evaluation inference configurations per benchmark.}
\label{tab:eval-protocol}
\end{table}

\paragraph{EnvScaler.}
We evaluate on a held-out set of $200$ tasks, randomly sampled from the full task pool and fully disjoint from the training split.

\paragraph{BFCL-v3.}
We use the official Berkeley Function Calling Leaderboard V3 codebase with four multi-turn subsets: base, missing-function, missing-parameter, and long-context. Models are served in function-calling (FC) mode. We report per-subset and average scores.

\paragraph{ACEBench.}
ACEBench evaluates $50$ tasks split into multi-step ($20$) and multi-turn ($30$) categories. A user simulator (\texttt{deepseek-v4-flash} via API) drives the multi-turn dialogues.

\paragraph{LiveCodeBench.}
We use the official \texttt{code\_generation\_lite} evaluation harness and report pass@1 on release v5 ($279$ tasks, Aug 2024--Feb 2025) and release v6 ($131$ tasks, Feb--May 2025).

\paragraph{EvalPlus.}
We use HumanEval+ v0.1.10 ($164$ tasks) and MBPP+ v0.2.0 ($378$ tasks) from the official EvalPlus codebase.

\subsection{Ablation and Analysis Setup}\label[appendix]{app:ablation-setup}

This section details the configuration of each analysis experiment in \Cref{sec:framework-analysis}.

\paragraph{Effect of Joint Optimization (\Cref{fig:composer-ablation}).}
Each row trains \ourmethod with a different margin $m$ and evaluates the resulting student on the EnvScaler test set without retrieval, the composer, or latent privileged context at inference, isolating the effect of joint optimization on the resulting policy. The ``Frozen $\phi_0$'' row trains with a frozen composer (no joint optimization). All jointly-optimized rows share the same training configuration except for $m$.

\paragraph{Training Dynamics (\Cref{fig:envscaler-training-dynamics}).}
Mean reward is periodically evaluated on the EnvScaler test set throughout training using \llmname{Qwen3-4B}.

\paragraph{Sensitivity Analysis (\Cref{fig:sensitivity-analysis}).}
(a)~\textit{Latent-token capacity}: we train separate \ourmethod variants with $K \in \{8, 16, 32, 64, 128\}$ and evaluate each resulting student alone on the EnvScaler test set ($n_\mathrm{ret}{=}3$ during training). (b--e)~\textit{Retrieval count}: we train separate \ourmethod variants with $K{=}32$, $m{=}0.05$, and $n_\mathrm{ret}\in\{1,\ldots,10\}$, then evaluate the resulting students on the EnvScaler test set and ACEBench without retrieval, the composer, or latent privileged context.

\paragraph{Behavioral Internalization (\Cref{tab:behavior-analysis}).}
\textit{Vanilla}: the unadapted \llmname{Qwen3-4B} backbone. \textit{Base + Composer}: the jointly optimized composer ($m{=}0.05$) paired with the unadapted backbone and latent privileged context ($n_\mathrm{ret}{=}3$, $K{=}32$). \textit{\ourmethod}: the distilled student policy evaluated without retrieval, the composer, or latent privileged context. All three are evaluated on the EnvScaler test set with otherwise identical inference settings.

\end{document}